\documentclass[10pt,twocolumn,letterpaper]{article}

\usepackage{iccv}
\usepackage{times}
\usepackage{epsfig}
\usepackage{graphicx}
\usepackage{amsmath}
\usepackage{amssymb}
\usepackage{textcomp}
\usepackage{bm}
\usepackage{mathrsfs}
\usepackage[dvipsnames]{xcolor}
\usepackage[normalem]{ulem}
\usepackage{outlines}
\usepackage{booktabs}
\usepackage{threeparttable}
\usepackage{tabularx}
\usepackage{multirow}
\usepackage{ifthen}
\usepackage{pifont}%
\newcommand{\cmark}{\ding{51}}%
\newcommand{\xmark}{\ding{55}}%

\DeclareMathOperator*{\argmin}{arg\,min}

\usepackage[pagebackref=true,breaklinks=true,letterpaper=true,colorlinks,bookmarks=false]{hyperref}

\renewcommand{\ie}{i.e.,} 
\newcolumntype{s}{>{\raggedleft\arraybackslash}X}

\newcommand{\ourmodel}{\textsc{LookOut}}
\newcommand{\ourdataset}{\textsc{ATG4D}}

\newcommand{\cutequationup}{\vspace*{-2pt}}
\newcommand{\cutsectionup}{\vspace*{-6pt}}
\newcommand{\cutsectiondown}{\vspace*{-2pt}}
\newcommand{\cutsubsectionup}{\vspace*{-5pt}}
\newcommand{\cutparagraphup}{\vspace*{-4pt}}

\newcommand{\Supp}{Supplementary Materials}
\newcommand{\supp}{supplementary materials}

\newboolean{arxiv}

\iccvfinalcopy %

\def\iccvPaperID{7104} %

\ificcvfinal\fi

\begin{document}

\title{LookOut: Diverse Multi-Future Prediction and Planning for Self-Driving}

\author{
   Alexander Cui \thanks{Denotes equal contribution}$~^{, 1, 3}$, Sergio Casas$~^{*, 1, 2}$, Abbas Sadat $^{*, 1}$, Renjie Liao$~^{1, 2}$, Raquel Urtasun$^{1, 2}$\\
   $\text{Uber ATG}^1$, $\text{University of Toronto}^2$, $\text{California Institute of Technology}^3$ \\
   {\small$\texttt{\{alex.yy.cui, abbas.sadat\}@gmail.com, \{sergio, rjliao, urtasun\}@cs.toronto.edu}$}
}

\maketitle
\ificcvfinal\thispagestyle{empty}\fi

\begin{abstract}

In this paper, we present \ourmodel{}, a novel autonomy system that perceives the environment, predicts a diverse set of futures of how the scene might unroll  and estimates the trajectory of the SDV by optimizing a set of contingency plans  over these future realizations.
In particular, we learn a diverse joint distribution over multi-agent future trajectories in a traffic scene that covers a wide range of future modes with high sample efficiency while leveraging the expressive power of generative models.
Unlike previous work in diverse motion forecasting, our diversity objective explicitly rewards sampling future scenarios that require distinct reactions from the self-driving vehicle for improved safety. 
Our contingency planner then finds comfortable and non-conservative 
 trajectories that ensure safe reactions to a wide range of future scenarios.
Through extensive evaluations, we show that our model demonstrates significantly more diverse and sample-efficient motion forecasting in a large-scale self-driving dataset as well as safer and less-conservative  motion plans in long-term closed-loop simulations when compared to current state-of-the-art models.

\end{abstract}

\cutsectionup
\section{Introduction}
\cutsectiondown
Self-driving vehicles (SDVs) have the potential to enhance considerably the safety of our roads as, unlike humans,  they can constantly scan the surrounding environment without getting distracted or being impaired while driving.
Key to the success of a self-driving vehicle is its ability to perceive its surroundings and predict the future trajectory of the traffic participants, %
particularly those that might affect its decision making. 
These predictions are then exploited  by the motion planning module to plan a safe and comfortable maneuver towards the goal.

\begin{figure}[t]
    \centering
    \vspace{-4pt}
    \includegraphics[width=\columnwidth]{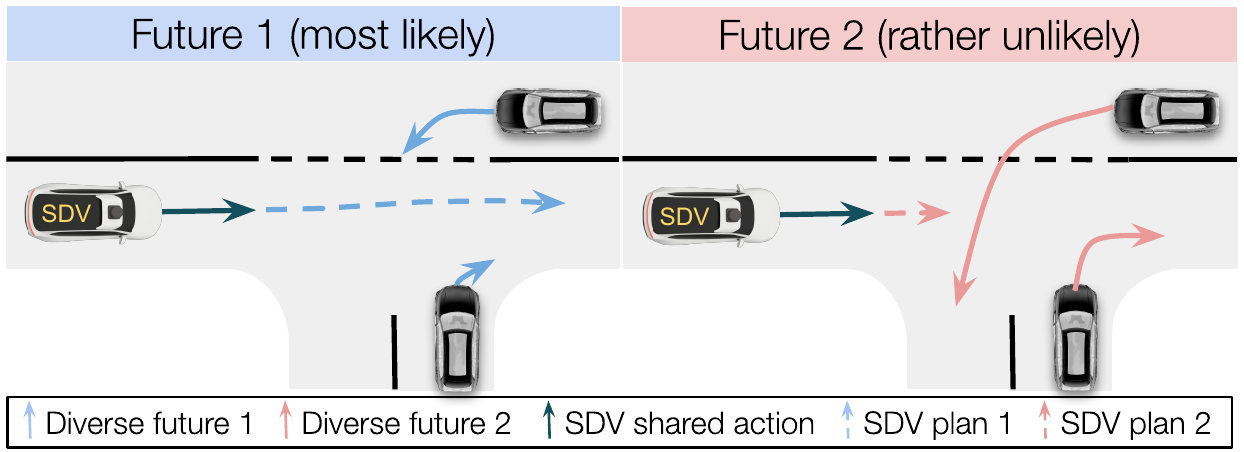}
    \caption{We illustrate the fact that the future is highly uncertain and multi-modal by showing 2 distinct futures at the scene-level.
    In such scenario, \ourmodel{} plans a short-term executable action that leads to 2 different contingent plans to stay safe in both cases.
    }
    \vspace{-8pt}
    \label{fig:motivating}
\end{figure}

Forecasting the behavior of traffic participants is very challenging as humans do not always follow the rules of the road and sometimes exhibit erratic behaviors. Furthermore, the scene might unroll in many possible ways in the future, depending heavily on the interactions between actors (e.g., at a merge, either actor A yields to actor B or vice versa).
While most works  predict each actor's future independently \cite{djuric2018motion, cui2018multimodal, phan2019covernet,chai2019multipath,casas2019spatially}, recent approaches model actor interactions and can produce samples that explain the full scene in a consistent manner \cite{alahi2016social,2019arXiv190501296R,tang2019multiple, casas2020implicit}. 
However, they require prohibitively large numbers of samples to cover the long-tails of the distribution. 
This is problematic since these long tails are critical for safety, as failing to take them into account might result in an accident (e.g., an impaired driver running a red traffic light perpendicularly to the SDV's intended trajectory). 
Thus, there is a need to develop prediction systems that can efficiently sample the diverse set of possible futures.
Unfortunatelly, existing approaches \cite{yuan2019diverse,yuan2020dlow} are not sample efficient as they trivially encourage diversity in euclidean space, thus utilizing samples to cover irrelevant actors or actions that do not impact the SDV's behavior.

Furthermore, existing motion planners cannot take advantage of  prediction systems that produce scene-consistent samples \cite{ziegler2014trajectory,fan2018auto, ajanovic2018search, sadat2019jointly}. Instead, they optimize the  expected cost by sampling the marginal distribution of each actor independently, thus ignoring the fact that some of these futures cannot happen at the same time (e.g., either the horizontal or vertical traffic can flow at a 4-way stop, but not both). These planners also assume that the SDV must commit to a single long-term trajectory, when in practice it can execute a short-term action and re-plan as newer sensor observations become available. 
As a consequence, they result in suboptimal and overly conservative trajectories \cite{zhan2016non,tas2018decision}, e.g., the SDV braking prematurely to react to an unlikely future instead of maintaining its speed as long as it is able to stop safely and comfortably later.

In this paper we propose  \ourmodel, a full end-to-end autonomy system that detects actors in the scene, predicts a diverse set of consistent futures with high sample efficiency, and plans an action that behaves defensively to potential hazards while not overreacting to
low probability dangers far into the future.
In particular, to address the sample inefficiency and limited mode coverage in motion forecasting we formulate this task as a \textit{diverse set prediction problem}, where each element in the set reflects one possible future at the scene level. 
To enable this set to cover the future modes that matter for our decision making, we directly optimize the diversity of the downstream ego-vehicle motion plans. 
Then, a scenario scoring module  estimates the probability of each future in the set, enabling our planner to account for unlikely but safety-critical scenarios without being overly conservative. 
Finally, we propose a novel contingency planner that is able to leverage  multiple consistent futures by planning separate long-term responses for each of future, while sharing an initial short-term action that behaves non-conservatively with respect to the futures and avoids immediate collision. 
Fig.~\ref{fig:motivating} shows an example of two diverse futures and the corresponding shared action and contingent plans.

We demonstrate the effectiveness of our approach in large-scale open-loop and closed-loop experiments that comprise a wide variety of complex scenarios. Our extensive experiments show that  \ourmodel{}'s driving is significantly safer as well as less conservative than previous state-of-the-art approaches. Furthermore,  there exists a trade-off between diversity and reconstruction quality of the forecasts; our approach can produce much better reconstruction than other methods with similar diversity and higher diversity  with similar reconstruction capability. %

\cutsectionup
\section{Related Work}
\cutsectiondown
The autonomy pipeline composed of cascading detection, motion forecasting and motion planning modules offers great advantages over black-box end-to-end models \cite{pomerleau1989alvinn,bojarski2016end,codevilla2018end,kendall2019learning,muller2018driving} such as safety, interpretability, error tracing, and data efficiency. 
Moreover, it has been recently shown that it can be learned end-to-end \cite{luo2018fast, casas2018intentnet, zeng2019end, sadat2020perceive, zeng2020dsdnet}.
Because of this, we focus our literature review on this approach.
For object detection, we simply leverage recent advances in 3D voxel-based object detection from LiDAR point clouds \cite{li20173d,engelcke2017vote3deep,yang2018pixor,zhou2018voxelnet,yang2018hdnet}, which have been shown to achieve great speed-accuracy tradeoffs.
In the following paragraphs, we dive deep into recent advances in motion forecasting and motion planning, given that the main contributions of our work reside on these modules.

\cutparagraphup
\paragraph{Motion Forecasting:}
A common approach for actor modeling has been to independently predict the trajectory of each actor \cite{rhinehart2018r2p2, djuric2018motion, casas2019spatially, chai2019multipath, jain2019discrete, phan2019covernet, zhao2020tnt}. These predictions can be represented as closed-form gaussian distributions \cite{djuric2018motion,casas2019spatially, chai2019multipath}, a classification or energy over a discrete grid/graph/set structure \cite{jain2019discrete,zhao2020tnt,phan2019covernet,zeng2020dsdnet},  or trajectory samples of a stochastic model \cite{rhinehart2018r2p2, Hong_2019_CVPR}.
One approach to tractably model the traffic multimodality jointly across actors is to stochastically sample one possible future scenario at a time, by sampling latent variables that encode the joint scene dynamics, and then decode the future trajectories \cite{2019arXiv190501296R,tang2019multiple,casas2020implicit}. These are mainly divided into autoregressive models \cite{2019arXiv190501296R,tang2019multiple}, and implicit latent variable models \cite{casas2020implicit}. However, these methods require a high number of samples to characterize the scene. In contrast, work in diverse motion forecasting has focused on achieving high sample-efficiency to cover the main modes of the distribution. This is especially important in self-driving as SDVs need to be able to anticipate rare or dangerous behavior by other actors on the road in order to plan safe responses. Recent work \cite{yuan2019diverse, yuan2020dlow} has explored how to encourage more diverse predictions from pretrained variational inference models \cite{sohn2015learning}. They train new encoders that output a fixed number of jointly diverse samples of latent variables. The formulation in \cite{yuan2019diverse} directly outputs the set of latent codes, and evaluates their diversity based on determinantal point processes (DPP). 
In the work of \cite{yuan2020dlow}, a set of multivariate gaussian distributions are sampled jointly via reparameterization trick with a shared noise, and a diversity loss based on the L2-distance between motion forecast samples is used to increase diversity in the predictions. 
Alternatively, \cite{huang2020diversitygan} trains a conditional GAN to output diverse samples using Farthest Point Sampling on the latent space to spread out over more modes of the latent space. Finally, \cite{park2020diverse} trains their trajectory samples to stay within the drivable area, allowing for greater diversity while retaining admissibility. While these works achieve greater diversity and accuracy in motion prediction, it is unclear how these improvements translate into better motion planning for autonomous agents.

\begin{figure*}[t]
    \vspace{-15pt}
    \centering
    \includegraphics[width=\textwidth]{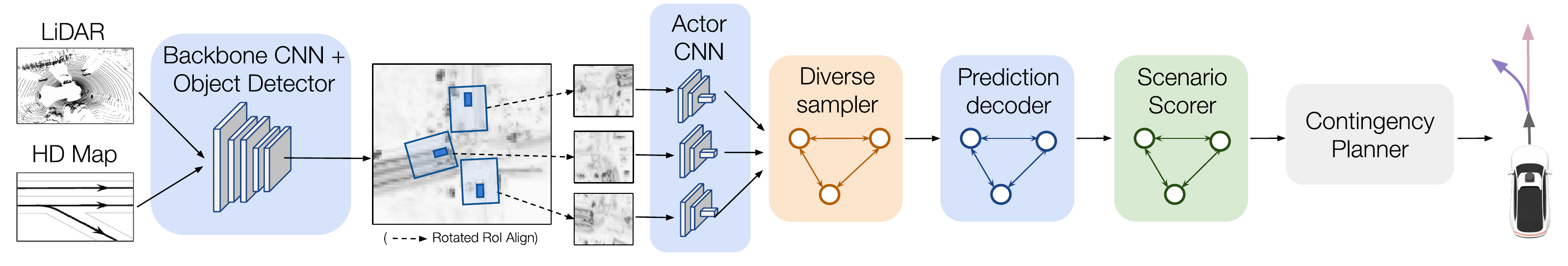}
    \caption{\ourmodel{} inference. For the learnable components, the colors denote different training stages. The {\color{RoyalBlue} backbone CNN, actor CNN and prediction decoder} are first trained (Section~\ref{sec:ilvm}), the {\color{Melon} diverse sampler} next (Section~\ref{sec:diverse_sampler}), and lastly the {\color{ForestGreen} scenario scorer} (Section~\ref{sec:scenario_scoring})}
    \vspace{-8pt}
    \label{fig:overview}
\end{figure*}

\cutparagraphup
\paragraph{Planning:}
In motion planning the goal is to generate a trajectory for the self-driving vehicle to drive safely, comfortably, and progressing toward the goal \cite{paden2016survey}. A popular approach to achieve this task is to design a cost function that encodes all the objectives above and find a minimum-cost trajectory. Such optimizations have been solved using continuous-optimization \cite{ziegler2014trajectory}, sampling \cite{sadat2019jointly, sadat2020perceive}, or search \cite{ajanovic2018search}. These methods achieve safety by including a collision cost in the objective function which is computed with respect to
the predicted trajectories of actors in the scene. However, in probabilistic settings where the predictions take the form of trajectory distributions, the above methods compute the collision cost in expectation, minimizing the expected cost over all future predicted scenarios with a single plan. 
Given a set of joint diverse predictions, it is possible that the SDV will need to plan for a greater number of unlikely, but safety-critical future scenarios that will require it drive defensively (e.g. yield, change lanes). However, it is undesirable to always brake preemptively for such rare scenarios, or ignore them altogether. Work in optimization-based motion planning \cite{yoon2008probabilistic, hardy2013contingency, zhan2016non, tas2018decision} plan for such rare scenarios by picking trajectory plans that ensure it can react to them safely, while also optimizing for objectives such as progress and comfort. \cite{hardy2013contingency} splits the planned trajectory into an initial shared section, and a set of branched plans that could be taken from the end of the shared section. 
\cite{zhan2016non, tas2018decision} predict the probability that a defensive maneuver will be necessary in the near future, and decide whether to postpone the decision to the future (where more information will be available). 

\cutsectionup
\section{Diverse Prediction and Planning}

\cutsectiondown
In this section, we break down the  autonomy problem of mapping sensor data to an executable action into several modules which provide  interpretability of the SDV decision making. 
Towards this goal, we first learn a joint perception and future prediction model that detects relevant objects  and estimates the joint distribution over all actors' future trajectories with an implicit latent variable model \cite{casas2020implicit} (Section~\ref{sec:ilvm}). Despite its sample inefficiency, such  generative model allows us to learn a very powerful and efficient trajectory decoder from latent samples.
Next, we leverage this decoder to learn a diverse latent sampler that achieves high sample efficiency from the planner's perspective (Section~\ref{sec:diverse_sampler}).
Then, we estimate the probability of each future realization in the set (Section~\ref{sec:scenario_scoring}).
Finally, we design a novel contingency planner that plans a safe trajectory for each possible future without being overly cautious (Section~\ref{sec:planning}). %
Fig.~\ref{fig:overview} depicts an overview of our approach. 

\cutsubsectionup
\subsection{Joint Perception and Motion Forecasting} \label{sec:ilvm}

In order to extract features useful for both detection and motion forecasting, we employ a convolutional \textit{backbone network} inspired by \cite{yang2018pixor, casas2018intentnet}, which takes as input  a history of voxelized LiDAR sweeps and a raster HD map, both in bird's eye view (BEV)  centered around the SDV. 
We then perform multi-class object detection with a shallow convolutional header to recognize the presence, BEV pose and dimensions of vehicles, pedestrians and bicyclists, and apply rotated RoI align \cite{ma2018arbitrary} to extract small feature crops from the scene context around each actor's location. Finally, an \textit{actor CNN} with max-pooling reduces the feature map of each actor $n$ into a feature vector, $x_n^{local}$. Since this local context lacks global information about the actor's pose with respect to the rest of the scene, we include the BEV centroid and rotation relative to the SDV $x_n^{global} = \{c_{x, n}, c_{y, n}, a_n\}$ as additional features, obtaining the final actor context $x_n = [x_n^{local}, x_n^{global}] \in \mathbb{R}^D$, where $[\cdot , \cdot]$ denotes channel-wise concatenation. We refer to the set of all the detected actors' contexts as $X = \{x_1, x_2, ..., x_N\}$.
The details about the LiDAR and map parameterization as well as the backbone network, object detector header, and actor CNN are left for the \supp{} as they are not the focus of our work and are highly inspired by previous literature \cite{luo2018fast,casas2018intentnet,casas2019spatially}.

 We parameterize the trajectory of each actor with a temporal series of the actor centroid in 2-dimensional Euclidean space, i.e., $y_n \in \mathbb{R}^{2T}$, where each trajectory is predicted in the actor's relative coordinate frame in Bird's Eye View (BEV) defined by its centroid and heading.
Our latent variable model then characterizes the joint distribution over actors' trajectories as follows:
\cutequationup
{\small
\begin{align}\label{eq:cvae_obj}
p(Y \vert X) = \int_Z p(Y \vert X, Z) p(Z|X) dZ, 
\end{align}%
}
where $Z = \{z_1, z_2, ..., z_N\}$ is a set of continuous latent variables that capture latent scene dynamics, and Y$ = \{y_1, y_2, ..., y_N\}$ is the future trajectories of all actors.
We assume a fixed prior $p(Z|X) \approx p(Z) = \prod_{n=1}^{N} p(z_n)$, where $z_n \sim \mathcal{N}(0, I) \in \mathbb{R}^L$.
Following \cite{casas2020implicit}, we adopt an implicit\footnote{``Implicit'' means $p(Y|X,Z)$ does not have analytical form.} decoder $Y = f_\theta(X, Z)$, where $f_\theta$ is a deterministic function parameterized by a spatially-aware Graph Neural Network (GNN) \cite{casas2019spatially}. 
Since from observational data we only obtain $(X, Y)$ pairs, a posterior or encoder function $q_\phi$ is introduced to approximate the true posterior distribution $p(Z \vert X, Y)$ during training \cite{sohn2015learning}, also parameterized by a GNN. This encoder function helps this model learn a powerful decoder, since  given only $X$ there could be many feasible $Y$ due to the inherent  multi-modality and uncertainty of the future.

The backbone network, detection header, actor CNN, encoder, and decoder are trained jointly for the tasks of object detection and motion forecasting.
We use binary cross-entropy with hard negative mining per class for the presence of an actor, and Huber loss for the regression targets (i.e., pose and dimension) \cite{yang2018pixor}. See the  \supp{}  for more details. We use the CVAE framework \cite{sohn2015learning} for the latent variable model, which optimizes the evidence lower bound (ELBO) of the log-likelihood $\log p(Y \vert X)$. 
Because the deterministic decoder leads to an implicit distribution over $Y$, we use Huber loss $\ell_{\delta}$ as the reconstruction loss \cite{casas2020implicit}, and reweight the KL term with $\beta$ as proposed by \cite{higgins2017beta}:
\cutequationup
{\small
\begin{align}
\mathcal{L}_{\text{forecast}} = & \frac{1}{NT} \sum_n^N \sum_t^T \ell_{\delta}(y_n^t - y_{n, GT}^t) \nonumber \\
& + \beta \cdot \text{KL}\left(q_{\phi}\left(Z \vert X, Y = Y_{GT} \right) \Vert p \left(Z \right ) \right),
\end{align}%
}
where the first term minimizes the reconstruction error between the trajectory samples $Y = \{y_n^t \vert \forall n, t\} = f_\theta(X, Z)$, $Z \sim q_{\phi}\left(Z \vert X, Y = Y_{GT} \right)$ and their corresponding ground-truth $Y_{GT}$, and the second term brings the privileged posterior $q_\phi(Z \vert X, Y = Y_{GT})$ and the prior $p(Z)$ closer. 

So far we have learned a powerful model of the future from which we can generate scene consistent samples for all actors in the scene. In particular, inference in this model works as follows: First, we encode the sensor data into actor contexts $X$. Then, we sample $K$ times from the prior $\{Z_k \sim p(Z) | \forall k\}$, and decode the scene latent samples deterministically in parallel to obtain each of the $K$ futures $\{Y_k =  f_\theta(X, Z_k) | \forall k\}$. Despite the high expressivity of this model and its attractive parallel sampling, it has two major drawbacks: (i) \textit{sample inefficiency}, and (ii) \textit{no closed-form likelihood}. In the following, we address (i) by learning to sample jointly a diverse set of latent codes that map into a covering distribution over trajectories and, (ii) by learning a categorical distribution over the diverse futures in the set.

\cutsubsectionup
\subsection{Planning-Centric Diverse Sampler} \label{sec:diverse_sampler}
\begin{figure}[t]
    \vspace{-10pt}
    \centering
    \includegraphics[width=\columnwidth]{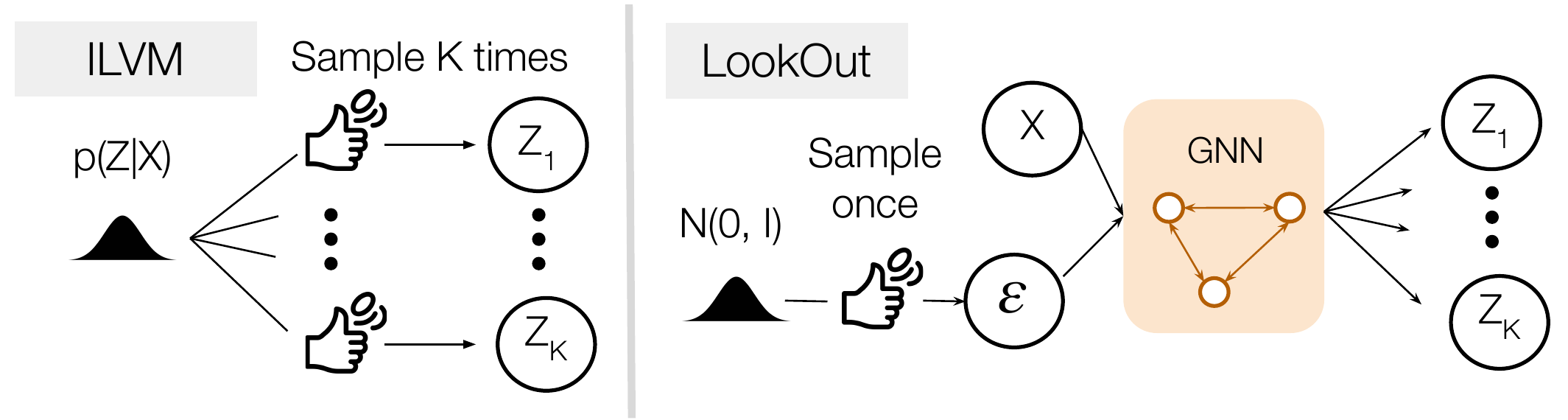}
    \caption{
		Obtaining $K$ latent samples from an implicit latent variable model (ILVM) implies sampling $K$ times independently from the prior. In contrast, our diverse sampler exploits a GNN mapping to predict $K$ latent samples from a single noise (in parallel).
	}
    \vspace{-8pt}
    \label{fig:sampling_contrast}
\end{figure}
The goal here is to remediate the sample inefficiency of the scene-level generative model presented in Section \ref{sec:ilvm} while exploiting its expressivity. To do so, we learn a \textit{diverse sampling function} $\mathcal{M}: X \mapsto \mathbf{Z}$ that maps the actor contexts $X$ coming from sensor data around each actor into a compact set of scene latent samples $\mathbf{Z} = \{Z_1, ..., Z_K\}$ whose decoded trajectories $\mathbf{Y}$ achieve good coverage. This sampler will then replace the Monte Carlo sampling from the prior distribution $p(Z)$ during inference, as illustrated in Fig.~\ref{fig:sampling_contrast}.

Since we want to leverage the decoder trained in Sec.~\ref{sec:ilvm}, which was trained to decode samples from a Gaussian approximate posterior, we would like the distribution over the set of latents induced by the diverse sampler to also be a Gaussian in order to reduce the distributional shift \cite{yuan2020dlow}. 
Thus, we assume $p(\mathbf{Z} \vert X) = \prod_{k=1}^{K} p(Z_k \vert X)$ where $p(Z_k \vert X) = \mathcal{N}(\mu_k, \Sigma_k)$, $\mu_k \in \mathbb{R}^{NL}$, and $\Sigma_k \in \mathbb{R}^{NL \times NL}$. 
To sample a set of latents $\mathbf{Z}$ that are distinct enough from each other such that they will be decoded into a set of diverse futures, we use the reparameterization trick \cite{kingma2013autoencoding} to map a shared noise $\varepsilon \sim \mathcal{N}(0, I) \in \mathbb{R}^{NL}$ across $K$ latent mappings $\{\mathcal{M}_{\eta_k} | k \in 1 \dots K\}$:
\cutequationup
{\small
\begin{align}\label{eq:diverse_sampler_params}
Z_k = \mathcal{M}_{\eta_k}(X, \varepsilon) = b_{\eta_k} (X) + A_{\eta_k}(X) \varepsilon,
\end{align}
}
where $\eta = \{\eta_k | \forall k \}$ is the set of learnable parameters, $\mu_k = b_{\eta_k} (X)$, and $\Sigma_k = A_{\eta_k}(X) A_{\eta_k}(X)^T$.

To handle the fact that the input $X \in \mathbb{R}^{ND}$  can vary in size (i.e.  the number of actors $N$ varies from scene to scene), we parameterize $\mathcal{M}$ with a pair of GNNs: one to generate the means and another to generate the covariances. Both GNNs assume a fully connected graph where each node is anchored to an actor, and initialize the node states as $\{x_n\}$. Then, we perform message passing to aggregate information over the whole scene at each node. Finally, each node in the first GNN predicts $a_n \in \mathbb{R}^{KL}$ via an MLP. Then, we can easily extract $A_{\eta_k}(X) = \text{diag}([a_1^{kL:(k+1)L}, \dots, a_N^{kL:(k+1)L} ])$. Similarly, each node in the second GNN predicts $b_n \in \mathbb{R}^{KL}$ via another MLP, and $b_{\eta_k}(X) = [b_1^{kL:(k+1)L}, \dots, b_N^{kL:(k+1)L} ]$.

The diverse latent codes $\mathbf{Z}$ can then be deterministically decoded via $Y_k = f_\theta(X, Z_k)$ with the decoder learned in Section~\ref{sec:ilvm}.
Through  sampling and decoding, we obtain a set of $K$ future trajectory realizations of all actors in the scene $\mathbf{Y} = \{Y_1, ..., Y_K\}$. This process   is parallel since it is performed by leveraging a pair of GNNs that perform all $K$ latent mapping in a single round of message passing $\mathbf{Z} \sim \mathcal{M}(X, \varepsilon; \eta)$. Then, we can batch the $K$ latent samples to decode them in parallel $\mathbf{Y} = f_\theta(\mathbf{Z}, X)$.

The objective of this diverse sampler is to be able to generate a set of futures $\mathbf{Y}$ that are diverse while recovering well the ground-truth observations $Y_{\text{gt}}$, which we can express through an energy $E(\mathbf{Y}, Y_{\text{gt}})$.
Moreover, to encourage minimal distribution shift to the inputs of the pretrained decoder $f_\theta$, we also minimize the KL divergence between all the diverse latent distributions $p(Z = Z_k \vert X)$ and the prior distribution $p(Z)$. In practice, this term makes the learning much more stable. To find the right balance between these two objectives, we add a hyperparameter $\beta$. Overall, the minimization can be formulated as:
\cutequationup
{\small
\begin{align}\label{eq:diverse_sampler_obj}
\min_\eta \quad E(\mathbf{Y}, Y_{\text{gt}}) + \beta \sum^{K}_{k=1} \text{KL} \left( p(Z_k \vert X) \Vert p(Z) \right),
\end{align}
}
where $\mathbf{Y} = \{Y_1, ..., Y_K\}$, $Y_k = f_{\theta}(X, Z_k)$, $Z_k = \mathcal{M}_{\eta_k}(X, \varepsilon)$ and the minimization is with respect to learnable parameters of the pair of GNNs $\eta$.
Note that the decoder is fixed, \ie{} $\theta$ is not optimized. 
This is key to maintain a high realism of the decoded trajectories, since the diversity objective can be otherwise cheated (e.g., by making other actors ``appear" right in front of the SDV).

Our energy function is composed of a few terms that promote the diversity while preserving data reconstruction: %
{\small
\begin{align}
E(\mathbf{Y}, Y_{\text{gt}}) = E_r(\mathbf{Y}, Y_{\text{gt}}) + E_p(\mathbf{Y}) + E_d(\mathbf{Y}).
\end{align}
}
We now define the energy terms in more details.
\cutparagraphup
\paragraph{Reconstruction Energy:} This term encourages that what happened in reality at the time the log was recorded to be captured by at least one sample:
\cutequationup
{\small
\begin{align}
	E_r(\mathbf{Y}) = \min_k \quad \ell_2(Y_k - Y_\text{gt}).
\end{align}
}
\vspace{-15pt}
\cutparagraphup
\paragraph{Planning Diversity Energy:}
We want to  increase prediction diversity in order to anticipate distinct future scenarios that require different SDV plans (e.g., a vehicle cuts in front of the SDV vs. keeps driving on its original lane). Thus, we  promote diverse samples that matter for the downstream task of motion planning by maximizing the following reward function:
\cutequationup
{\small
\begin{align}
R(\mathbf{Y}) = \frac{1}{K} \sum^K_{i=1} \sum^K_{j \neq i} \ell_2(\tau_i - \tau_j),
\end{align}
}
where $\tau_i = \tau(Y_i)$ refers to the SDV trajectory planned for predicted scene sample $Y_i$ by our contingency motion planner outlined in Section~\ref{sec:planning}. Since the optimal planned trajectory for each scene $\tau_i$ is not differentiable with respect to $Y_i$, we leverage the REINFORCE gradient estimator to express the energy $E_p$ as a function of the log-likelihood under the diverse sampler
\cutequationup
{\small
\begin{align}
E_p(\mathbf{Y}) &= - \mathbb{E}_{\mathbf{Y}} [R(\mathbf{Y})] \approx	- \log p(\mathbf{Z} \vert X) R(\mathbf{Y}) \\
&=\frac{1}{K (K-1)} \sum^K_{i = 1} \sum^K_{j\neq i} - \log p(Z_i, Z_j) \ell_2(\tau_i - \tau_j) \notag,
\end{align}
}
where $\log p(Z_i, Z_j) = \log p(Z_j) + \log p(Z_i)$. The approximation comes from a Monte Carlo estimation of the marginalization over $\mathbf{Z}$.

\cutparagraphup
\paragraph{General Diversity Energy:}
Since the signal from the planning-based diversity can be sparse for scenes that do not have any actors interacting with the SDV, we additionally encourage diversity in the behaviors of all actors: %
\cutequationup
{\small
\begin{align}
E_d(\mathbf{Y}) = \frac{1}{K(K-1)} \sum^K_{i=1} \sum^K_{j \neq i} \text{exp}(-\frac{\ell_2(Y_i - Y_j)}{\sigma_d}).
\end{align}
}
With our proposed diverse sampler, each $\mathbf{Y}$ induced by a different noise $\varepsilon$ efficiently covers well the distribution over futures. Thus, during inference we can simply take the set induced by the mode $\varepsilon=0$ to eliminate all randomness.
We note that determinism is important in self-driving  for safety, verification, and reproducibility.

\subsection{Scenario Probability Estimation} \label{sec:scenario_scoring}

The diverse set of $K$ future realizations $\mathbf{Y} = \{Y_1, ..., Y_K\}$ provides the coverage needed for safe motion planning. However, for  accurate risk assessment we need to estimate the probability distribution over each future realization in the set.
To achieve this goal, we augment our model to also output a score for all future realizations $l = s_\psi(X, \mathbf{Y})$, where $s_\psi$ is a GNN that takes as input the actor features and all $K$ sample future scenarios. We can then easily recover a distribution over such scores by re-normalization. 
Thus, the  probability of each sample is 
\cutequationup
{\small
\begin{align}\label{eq:sample_probability_score}
    p_\psi(Y_k \vert X) = \frac{\exp(l_k)}{\sum_{k'} \exp(l_{k'})}.
\end{align}
}
Since we only have access to a single ground truth realization (i.e., the one that occur in the training log),  we train the scoring function $s_\psi$ to match the approximate categorical distribution over future scenarios $q(Y_k \vert X)$ under the $\text{KL}(p_\psi \Vert q)$ divergence. We define this approximate distribution as follows:
\cutequationup
{\small
\begin{align}\label{eq:approx_probability_score}
    q(Y_k \vert X) = \frac{\exp(- \alpha \ell_2(Y_k - Y_{GT}))}{\sum_{k'} \exp(- \alpha \ell_2(Y_k' - Y_{GT}))},
\end{align}
}
where $\alpha=10$ is a temperature hyperparameter we chose empirically.

\subsection{Contingency Planner} \label{sec:planning}
The goal of the motion planning module is to generate safe, comfortable and not overly conservative trajectories for the SDV to execute. We achieve this  through \textit{Model Predictive Control}, where a trajectory is planned considering a finite horizon, and is executed until a new trajectory is replanned upon availability of a new LiDAR sweep.  Most planning frameworks in the literature \cite{paden2016survey, gu2012road,ajanovic2018search,sadat2019jointly} take an optimization-based  approach where the trajectory that minimizes the expected  cost  is selected for execution:
\cutequationup
{\small
\begin{align}
    \label{eq:plt}
    \tau^*_{0:T} = \argmin_{\tau_{0:T} \in \mathcal{T}_{0:T}(\textbf{x}_0)} \mathop{\mathbb{E}}_{p(Y)} c(\tau_{0:T}, Y),
\end{align}
}
where 
$\mathcal{T}_{0:T}(\textbf{x}_0)$  denotes the set of possible trajectories starting from SDV state $\textbf{x}_0$ up to the horizon $T$, and $c$ denotes the planner cost function. Note that the expectation is over the distribution of possible future realizations of all actors  $P(Y)$. However, the above formulation does not exploit the fact that only one of the predicted scenarios will happen in the future and is conversely optimizing for a single trajectory that is "good" in expectation. Note that if we  change the expectation in Eq.~\ref{eq:plt} to the \textit{max} operator, the planner will optimize for the worst-case scenario regardless of its likelihood. Consequently the planner will become over-conservative, e.g.,~it will apply a hard-break for a very low probability scenario where a vehicle crosses SDV lane, as shown in \cite{zhan2016non, tas2018decision}.

\begin{figure}[t]
    \centering
    \includegraphics[width=0.8\columnwidth]{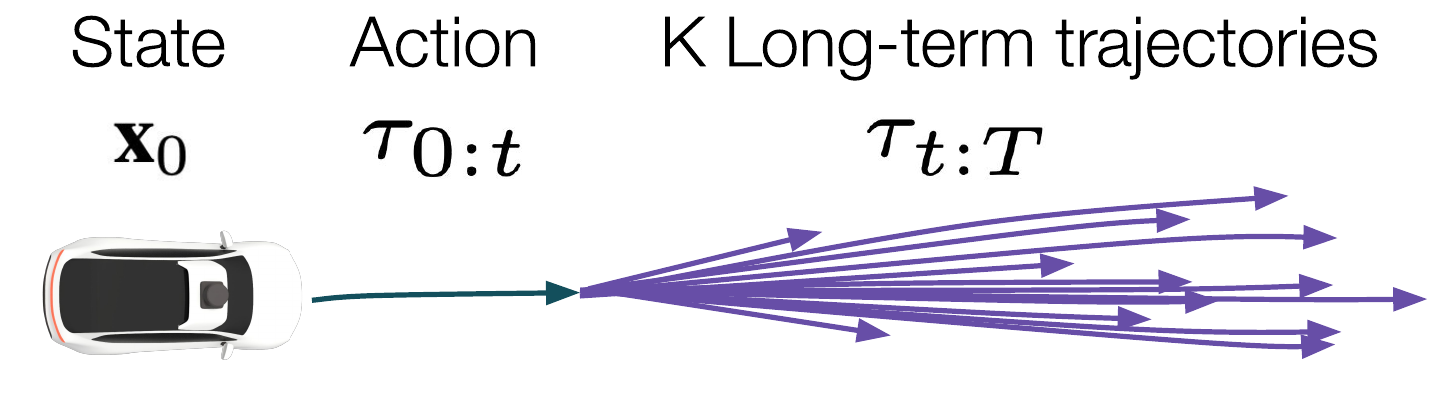}
    \caption{\textbf{Contingency planning paradigm}. The cost-to-go of a short term ego-action is captured by the ability to react to the $K$ diverse predicted futures with the $K$ most suitable ego-trajectories.
    }
    \vspace{-10pt}
    \label{fig:cartoon_contingent_plans}
\end{figure}

In this paper, we take a different approach where instead of finding a single motion plan for multiple futures, we generate a single common immediate action, followed by a set of future trajectories, one for each  future realization of the scene, as shown in Fig.~\ref{fig:cartoon_contingent_plans}. 
This \textit{contingency planning} paradigm finds an immediate action $\tau_{0:t}$ that is safe with respect to all the possible realizations in $Y$ and comfortably bridges into a set of contingent trajectories, where each is specifically planned for a single future realization. Such decision-postponing avoids over-conservative behaviors while staying safe until more information is obtained. Importantly, the described safe motion planning is only possible if the set of predicted future scenarios is diverse, and covers possible realizations, including low likelihood events.

Specifically, we plan a short-term trajectory that is safe with respect to all possible futures and allows a proper contingent plan for each future realization:
\cutequationup
{\small
\begin{align}
    \label{eq:contingency}
    \tau^*_{0:t} = \argmin_{\tau_{0:t} \in \mathcal{T}_{0:t}(\textbf{x}_0)}\bigg(\overbrace{\max_Y c(\tau_{0:t}, Y)}^{\text{action cost}} + \overbrace{\sum_{Y_i \in \mathbf{Y}} p(Y_i)g(\textbf{x}_t, Y_i)}^{\text{cost-to-go}}\bigg)
\end{align}
}
where $g(\textbf{x}, Y) = \min_{\tau_{t:T} \in \mathcal{T}_{t:T}(\textbf{x})} c(\tau_{t:T}, Y)$ represents the minimum cost trajectory from time $t$ to $T$ starting from the state $\textbf{x}$ and assuming a single future realization $Y$. 

\begin{table*}[t]
  \vspace{-20pt}
  \small
	\centering
	\begin{threeparttable}
        \begin{tabularx}{\textwidth}{
                        l |  %
                        s s s s s s s s%
                        }
        \toprule
                Model &
                CR$(\%)$  &
                $\frac{\text{Progress}}{collision}(m)$& 
                Progress$(m)$& 
                Jerk$(\frac{m}{s^3})$  
                & Lat.Acc.$(\frac{m}{s^2})$ 
                & Acc$(\frac{m}{s^2})$ 
                & Decel$(\frac{m}{s^2})$  \\
            \midrule
CVAE-DPP\cite{yuan2019diverse}  & 17.07 & 123.97 & 21.17 & 11.99 & \textbf{0.06} & 1.11 & 0.80 \\
CVAE-DLow\cite{yuan2020dlow}  & 14.63 & 377.07 & 55.18 & 5.22 & 0.15 & 0.84 & 0.54 \\
MultiPath\cite{chai2019multipath}   & 12.20 & 394.37 & 48.09 & 12.92 & 0.13 & 1.24 & 0.80 \\
CVAE\cite{sohn2015learning}  & 8.54 & 655.22 & 55.93 & 7.22 & 0.15 & 0.96 & 0.62 \\
ESP\cite{2019arXiv190501296R}  & 11.59 & 464.44 & 53.81 & 6.52 & 0.15 & 0.89 & 0.57 \\
ILVM\cite{casas2020implicit}  & 10.98 & 553.96 & 60.80 & 5.50 & 0.16 & 0.86 & 0.56 \\
\midrule
\ourmodel & \textbf{7.93} & \textbf{790.37} & \textbf{62.65} & \textbf{4.69} & 0.37 & \textbf{0.79} & \textbf{0.53} \\
	  \bottomrule
		\end{tabularx}
    \end{threeparttable}
    \caption{\textbf{End-to-end driving results in closed-loop simulation}. All motion forecasting baselines use the PLT planner \cite{sadat2019jointly} (Eq.~\ref{eq:plt}) as they don't propose a motion planner. Please see our \supp{} for results when they are paired with our planner (Eq.~\ref{eq:contingency}).}
    \vspace{-5pt}
	\label{table:main_closed_loop}
\end{table*}

\cutparagraphup
\paragraph{Cost Function:}
The planner cost function $c(\cdot) = \sum_i w_i s_i(\cdot)$ is a linear combination of various carefully crafted subcosts $s_i$ that encode different aspects of driving including safety, comfort, traffic-rules and the route. 
Here, $w = \{w_i | \forall i \}$ is a set of learnable parameters. However, learning these parameters in the contingency planning paradigm (Eq.~\ref{eq:contingency}) is an open problem since we only have expert demonstrations for the future that occured at the time of the log. Thus, we leave this for future work, and leverage the weights learned through Eq.~\ref{eq:plt} by \cite{sadat2019jointly}.
Regarding the subcosts, collision and safety-distance subcosts penalize SDV trajectories that overlap with the predicted trajectories of other actors or have high speed in close distance to them. Similarly, trajectories that violate a headway buffer to the lead vehicle are penalized. Other subcosts promote driving within the lane and road boundaries, and penalize trajectories that go above speed-limit or violate a red-traffic light. Finally, motion jerk, high forward acceleration, deceleration, and lateral acceleration of the trajectories are penalized to promote comfortable maneuvers. The details of all the subcosts can be found in the \supp. 

\cutparagraphup
\paragraph{Inference:}
We take a sampling approach to solve the minimization in Eq.~\ref{eq:contingency}. Specifically, we generate a set of pairs $\{(\tau_{0:t}, \mathcal{T}_{t:T}(\tau_{t}))\}$, which include possible short-term trajectories $\tau_{0:t}$ and their possible subsequent set of trajectories $\mathcal{T}_{t:T}(\tau_{t}))$. It is important to consider a dense set of initial actions such that the final executed trajectory is smooth and comfortable. Similarly, a dense set of long-term trajectories enables the planner to find a proper contingent plan for the future and thus obtain a more accurate cost-to-go for the initial action.
In order to manage the complexity of the search space above, we take the following sampling strategy: (i) first a set of (spatial) paths are generated, (ii) for each path, a set of initial velocity profiles are sampled, creating the set of short-term trajectories, (iii) conditioned on the end state of these initial trajectories, another set of velocity profiles are sampled for the rest of the planning horizon assuming the SDV follows the same path. In total, the sample set contains $\approx$ 240 actions and for each action there are $\approx$ 260 long-term trajectories.
The above path and velocity generation are done in Frenet-frame of the desired lane center line, by sampling lateral and longitudinal profiles \cite{werling2010optimal, sadat2019jointly}. For more details see  the \supp.

\cutsectionup
\section{Experiments}
\cutsectiondown
In this section we describe our experimental setup, followed by the results and discussions.

\cutsubsectionup
\subsection{Experimental Setup}
\cutparagraphup
\paragraph{Dataset:} \ourdataset{} \cite{yang2018pixor} is composed of over one million frames of LiDAR, HD maps with very accurate object tracks. It was collected with careful expert drivers in several North American cities. All models are trained to predict 5-second trajectories, given 1 second of LiDAR history. We evaluate motion forecasting in the test set of this dataset.

\cutparagraphup
\paragraph{Closed-loop simulator:} We use a simulated LiDAR environment \cite{manivasagam2020lidarsim} for closed-loop experiments where we evaluate the quality of our end-to-end
driving model, recreated from real static environments and actors. These scenarios are curated from real driving logs to be particularly challenging, and they do not overlap with those in \ourdataset{} in order to evaluate generalization. When replaying the scenario, the actors switch to reactive actors \cite{Treiber_2000} if the scenario diverges from the original one due to SDV actions. The simulation is unrolled for $\sim$18 seconds at intervals of 100 milliseconds, which is the same time it takes to acquire a new LiDAR sweep in the data collection vehicle.
We note that all training happens on real offline data, but it transfers well to the simulated environment due to its high realism. 

\begin{figure*}[t]
    \centering
    \begin{tabular} {c@{\hspace{.5em}}c@{\hspace{.5em}}c@{\hspace{.5em}}c}
        \raisebox{-0.5\height}{\includegraphics[width=0.23\linewidth]{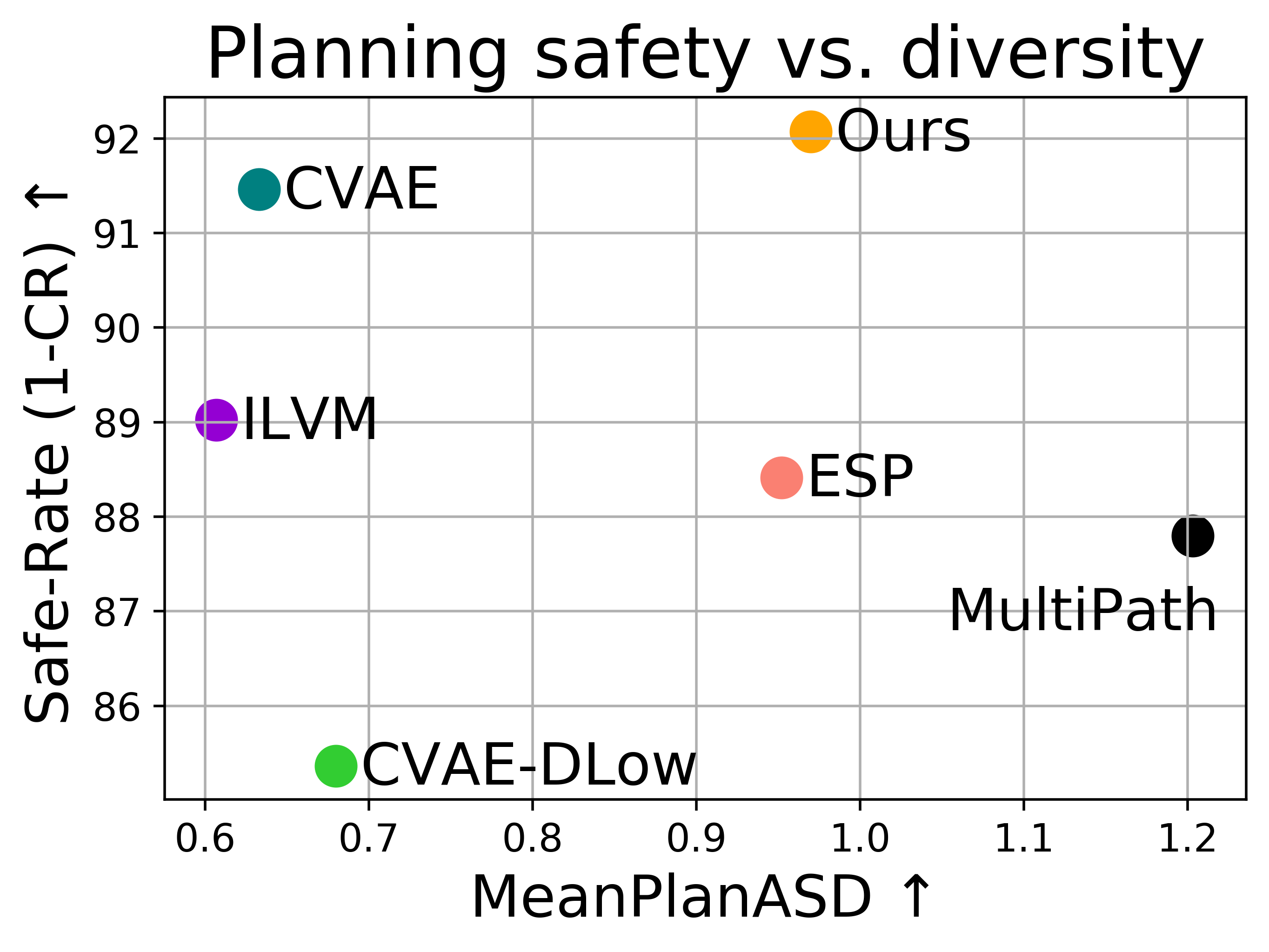}} &
        \raisebox{-0.5\height}{\includegraphics[width=0.23\linewidth]{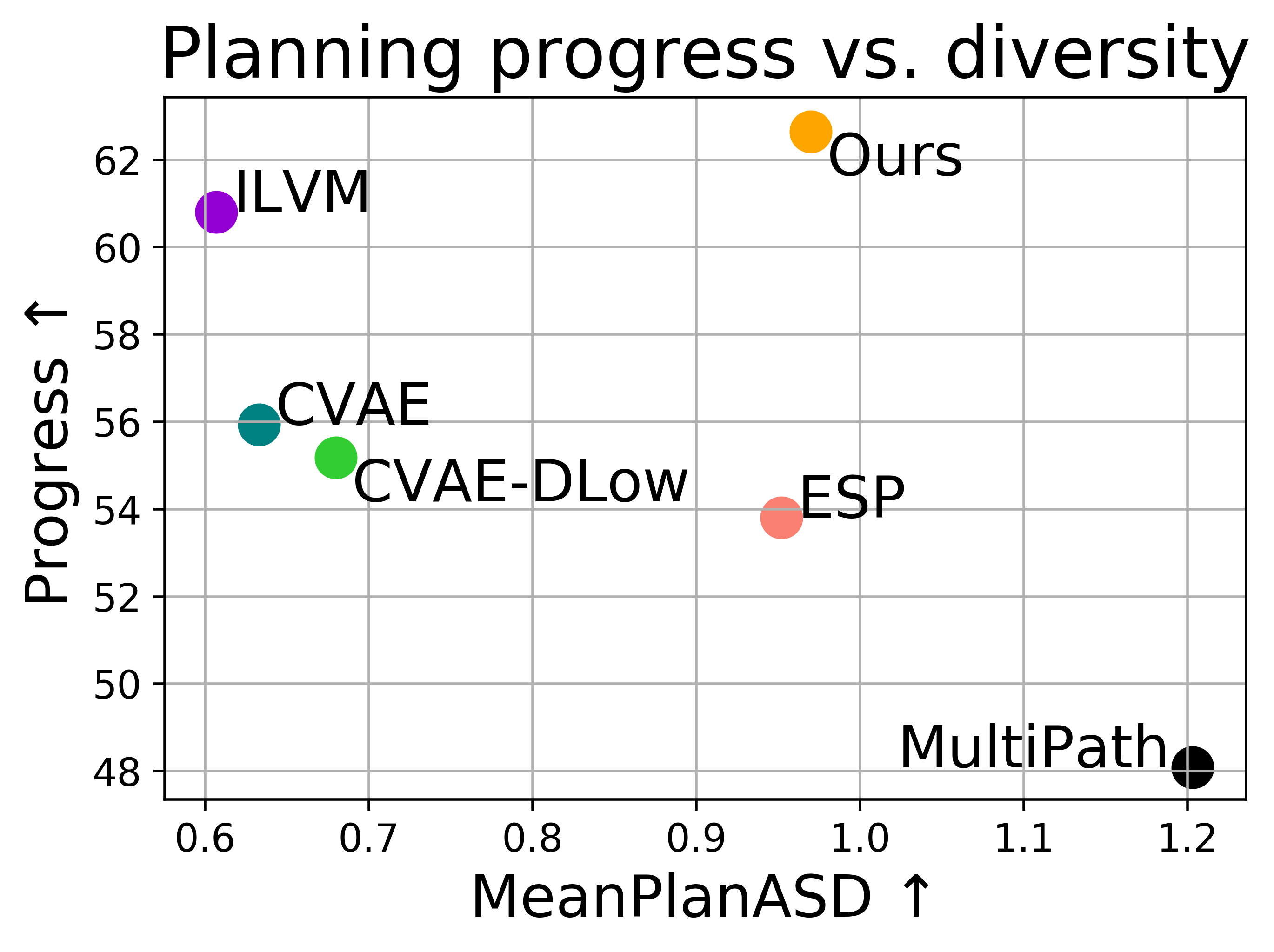}} &
        \raisebox{-0.5\height}{\includegraphics[width=0.25
        \linewidth]{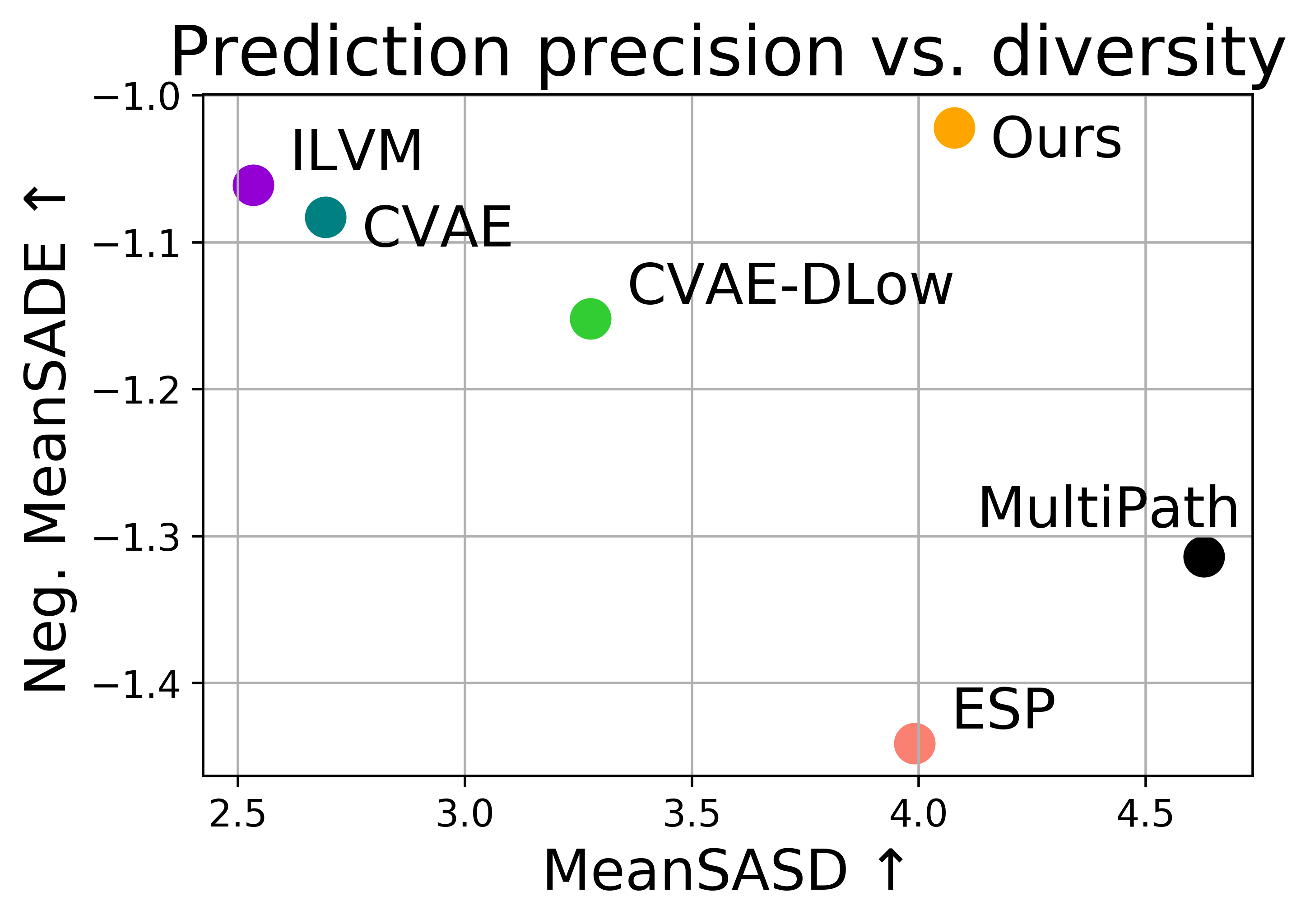}} &
        \raisebox{-0.5\height}{\includegraphics[width=0.25\linewidth]{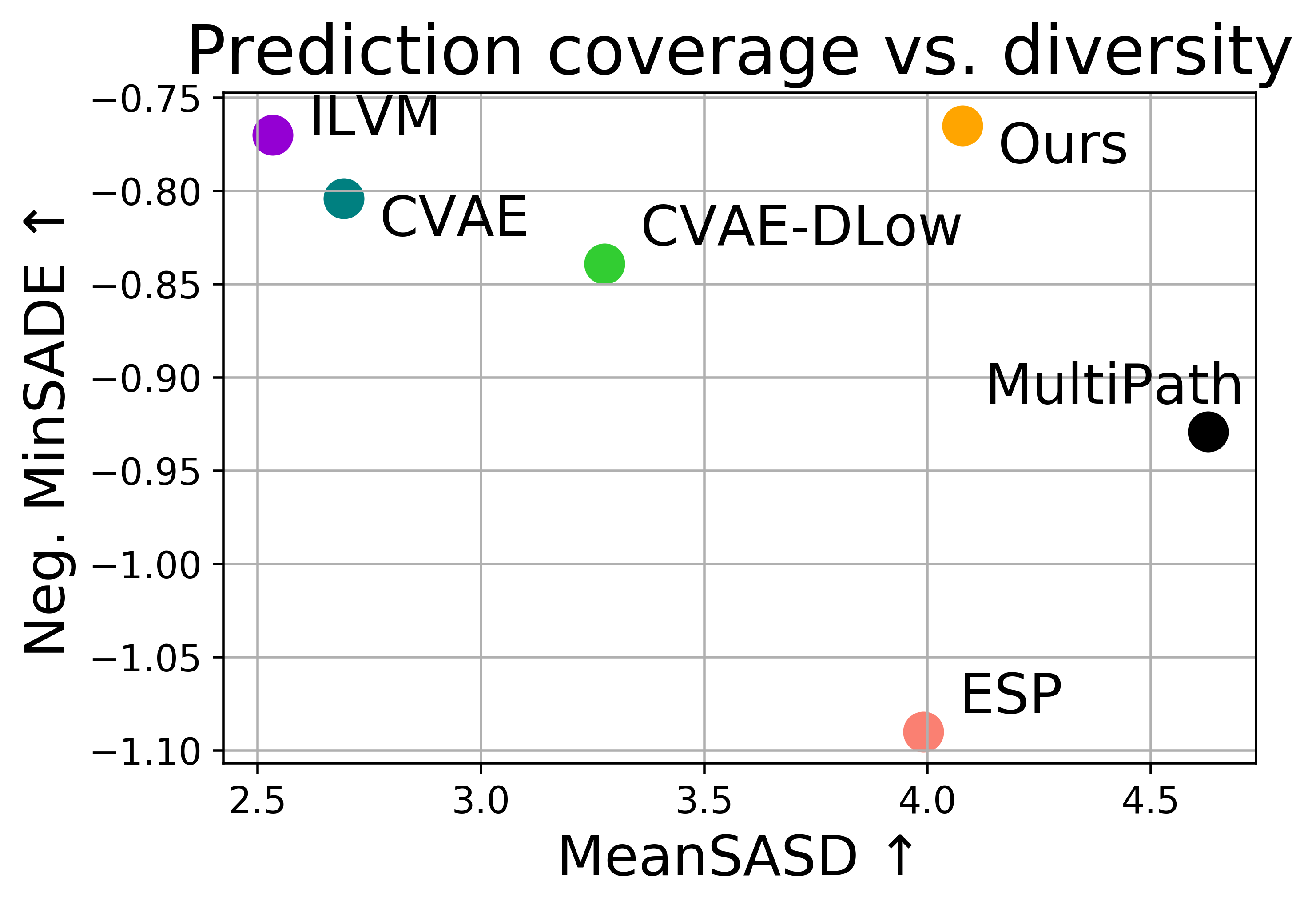}} \vspace{.5em} \\
    \end{tabular}
    \vspace{-10pt}
    \caption{
        \textbf{Planning quality and prediction reconstruction as a function of diversity}. More diversity is not always better. We do not include CVAE-DPP in these visualizations for clarity, as it has much lower performance than other models and would be off-the-charts. 
    }
    \vspace{-10pt}
    \label{fig:diversity_plots}
\end{figure*}

\begin{table*}[t]
\small
\centering
\begin{threeparttable}
\setlength\tabcolsep{5pt} %
\begin{tabularx}{\textwidth}{
                l |
                c
                c
                c
                c |
                r r r r r r r r%
                }
                \toprule
        
        ID &
        $\mathcal{M}_\eta$ &
        $E_p$ &
        $s_\psi$ &
        Planner &
        CR$(\%)$  &
        $\frac{\text{Progress}}{collision}(m)$&
        Progress$(m)$&
        Jerk$(\frac{m}{s^3})$
        & Lat.Acc.$(\frac{m}{s^2})$
        & Acc$(\frac{m}{s^2})$
        & Decel$(\frac{m}{s^2})$ \\

        \midrule

        $M_1 $ & \xmark & {\scriptsize N/A} & \xmark & {\scriptsize Conting.} & 9.15 & 709.60 & 64.90 & \textbf{4.40} & 0.38 & \textbf{0.77} & \textbf{0.52} \\
        $M_2 $ & \cmark & \xmark & \cmark & {\scriptsize Conting.}  & 10.98 & 626.79 & \textbf{68.79} & 8.77 & 0.20 & 1.09 & 0.73 \\
        $M_3 $ & \cmark & \cmark & \xmark & {\scriptsize Conting.} & 9.15 & 658.58 & 60.24 & 4.96 & 0.35 & 0.79 & 0.53 \\
        $M_4 $ & \cmark & \cmark & \cmark & {\scriptsize PLT} & 12.80 & 436.29 & 55.87 & 6.26 & \textbf{0.16} & 0.90 & 0.59 \\

        \midrule
        $\ourmodel$ & \cmark & \cmark & \cmark & {\scriptsize Conting.} & \textbf{7.93} & \textbf{790.37} & 62.65 & 4.69 & 0.37 & 0.79 & 0.53 \\

        \bottomrule
\end{tabularx}
\end{threeparttable}
\caption{\textbf{Ablation study} on the effect of the diverse sampler $\mathcal{M}_\eta$, planning diversity energy $E_p$, scenario scorer $s_\psi$ and motion planner towards the end-to-end driving capability (evaluated in closed-loop simulations).}
\vspace{-5pt}
\label{table:ablation_closed_loop}
\end{table*}

\begin{figure*}[t]
    \centering
    \begin{tabular}{c@{\hspace{.32em}}c@{\hspace{.64em}}c}
        \textbf{Scenario 1} &
        \textbf{Scenario 2} &
        \textbf{Scenario 3} \\
        \raisebox{-0.5\height}{\includegraphics[trim={.5cm .33cm .5cm .33cm},clip, width=0.32\textwidth]{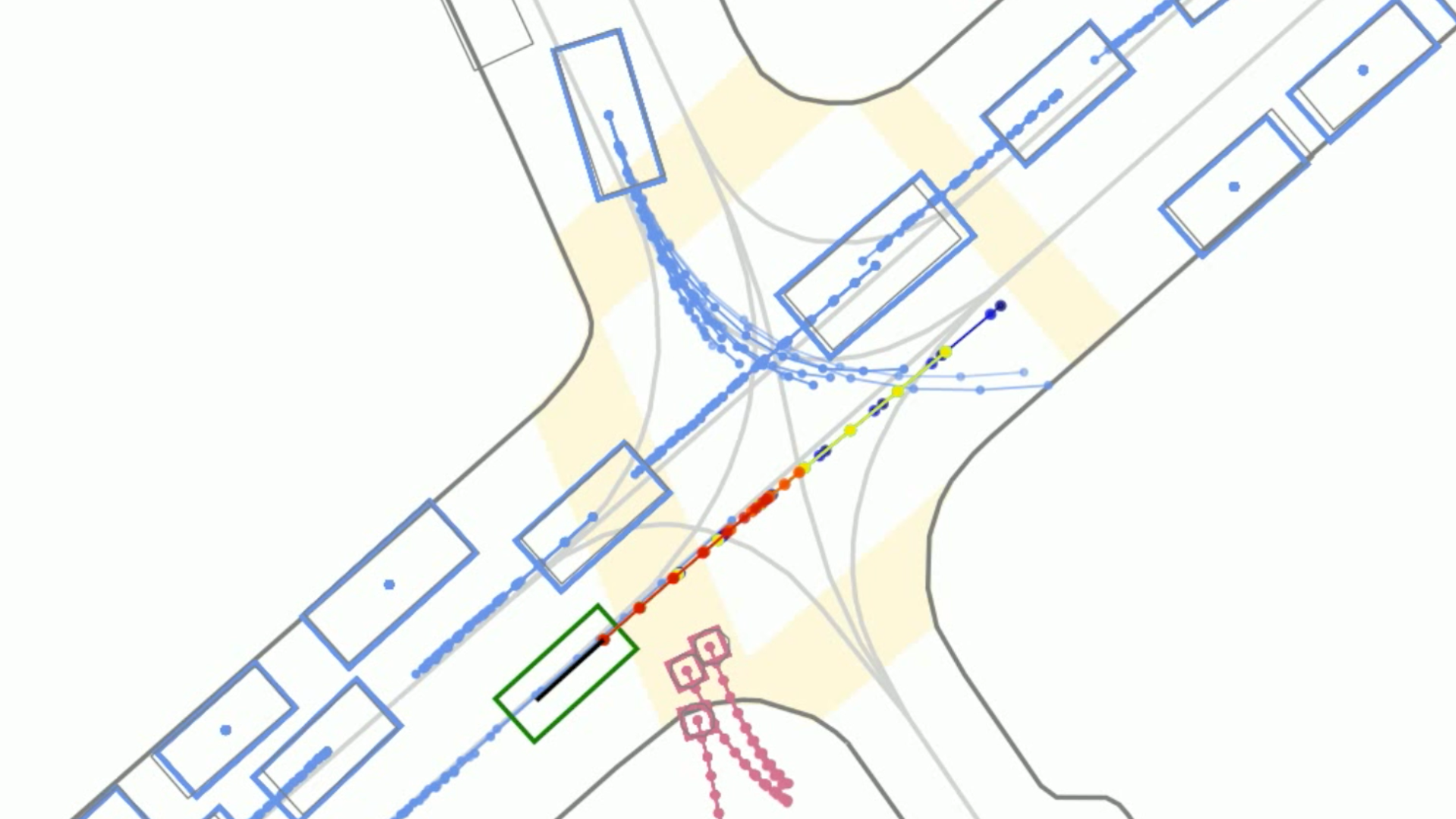}} &
        \raisebox{-0.5\height}{\includegraphics[trim={.5cm .33cm .5cm .33cm},clip, width=0.32\textwidth]{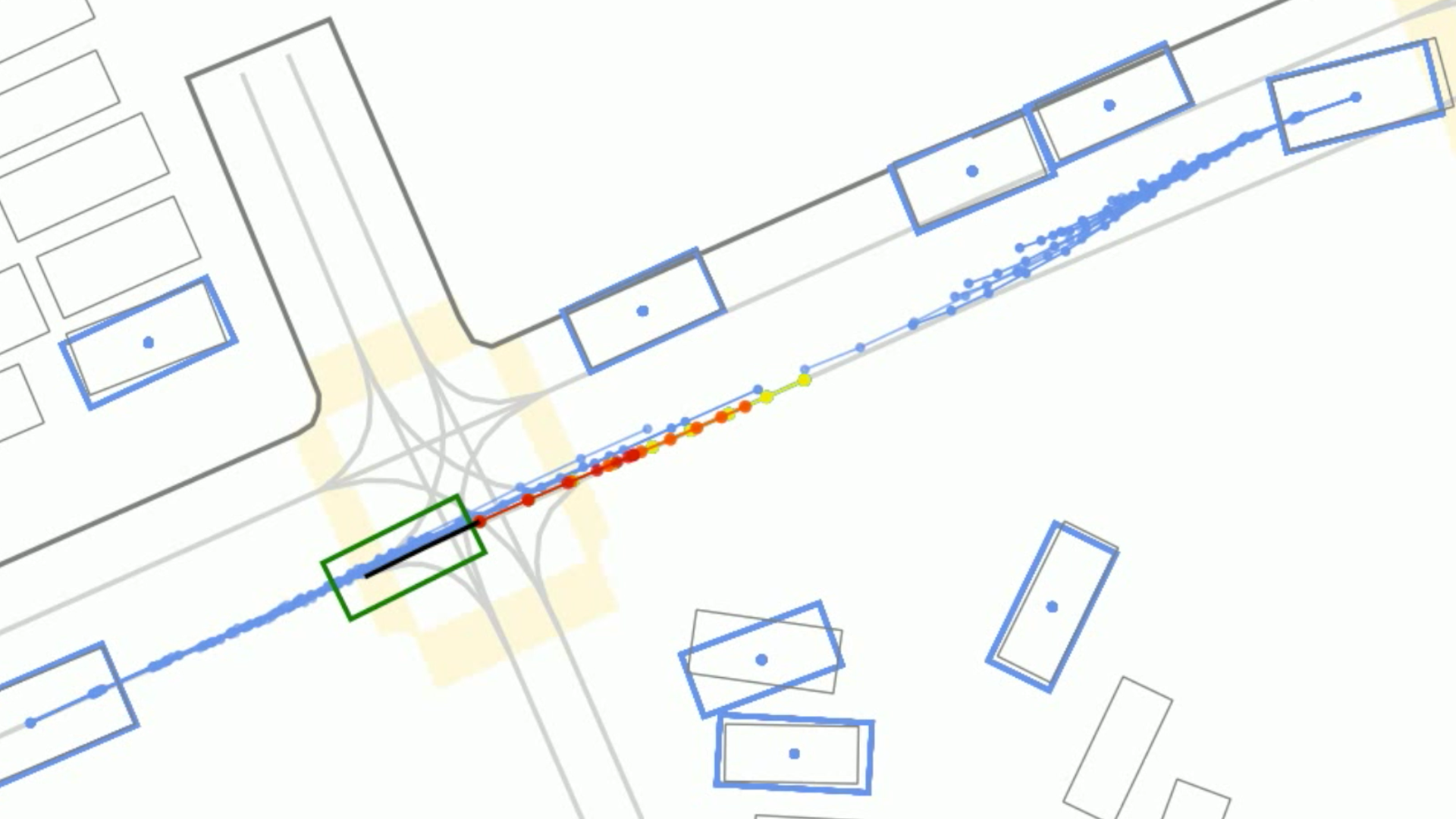}} &
        \raisebox{-0.5\height}{\includegraphics[trim={.5cm .33cm .5cm .33cm},clip, width=0.32\textwidth]{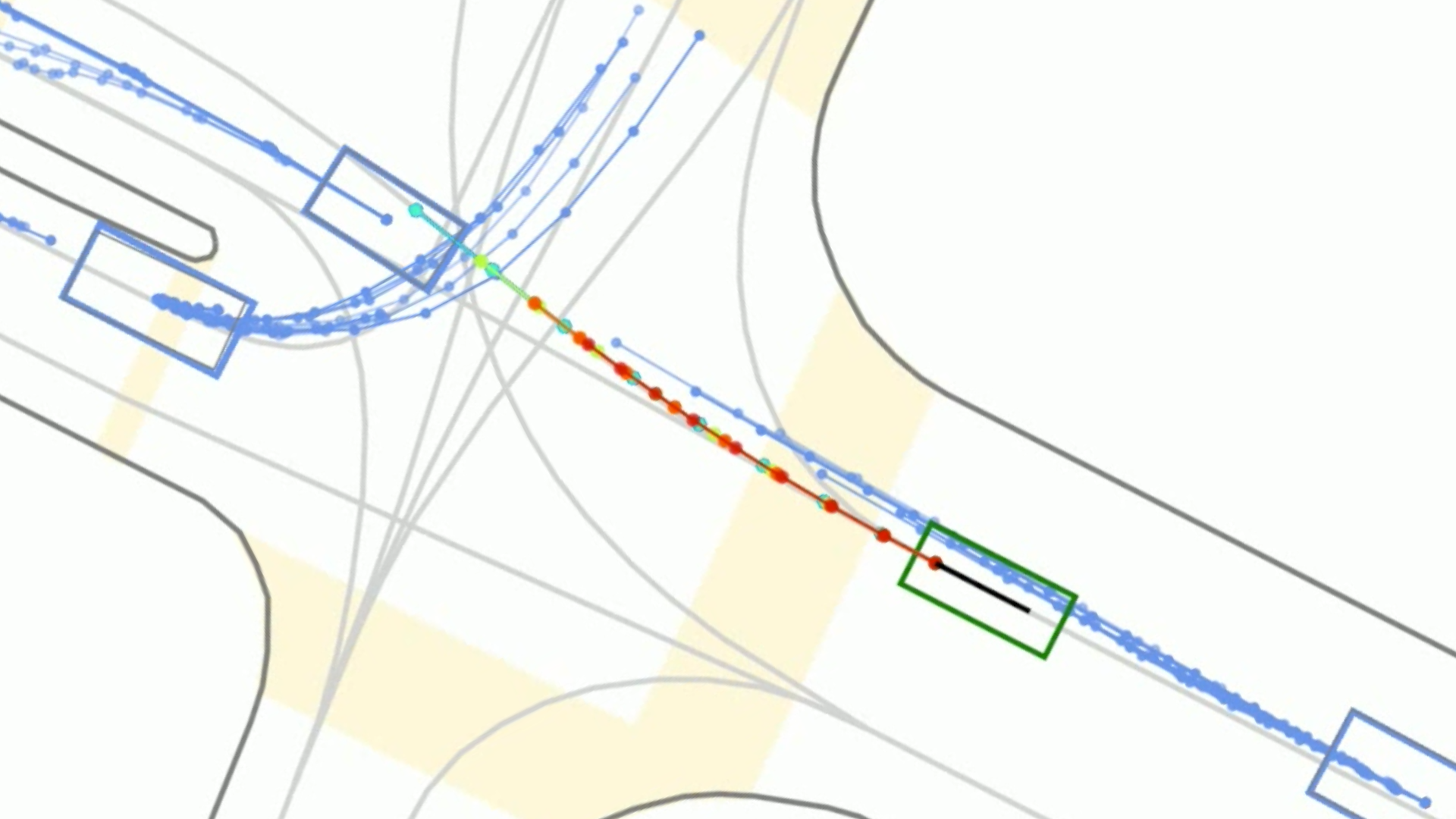}} \\
    \end{tabular}
    \caption{
		\textbf{Diverse multi-future predictions and plans} in closed-loop, zoomed in.
        Object detections and motion forecasts are blue for vehicles and pink for pedestrians.
        The green bounding box is the SDV, its immediate action (1s) is shown in black (starting from its rear axle), and its contingent trajectories planned for each possible future scenario are shown in distinct colors.
        LiDAR points are not visualized.
        }
    \vspace{-10pt}
    \label{fig:qualitative}
\end{figure*}

\cutparagraphup
\paragraph{Baselines:} For motion forecasting, we use state-of-the-art baselines in multi-modal and diverse prediction, all of them trained end-to-end with the same backbone network and object detector architectures for a fair comparison, following the experimental setup in \cite{casas2019spatially, casas2020implicit}.
\textit{MultiPath} \cite{chai2019multipath}, \textit{CVAE} \cite{sohn2015learning},  \textit{CVAE-DPP} \cite{yuan2019diverse} and \textit{CVAE-DLow} \cite{yuan2020dlow} model the distribution over each actor's future trajectories independently. Thus, to construct a scene sample for these baselines we sample a random trajectory for each actor, following \cite{casas2020implicit}.
For approaches that model the joint distribution over all actors' future trajectories, we benchmark  against  \textit{ESP} \cite{2019arXiv190501296R} and \textit{ILVM} \cite{casas2020implicit}. 
To compare \ourmodel{} to the baselines in the motion planning task, we use the state-of-the-art PLT planner \cite{sadat2019jointly} (Eq.~\ref{eq:plt}) for those motion forecasting models that did not propose a planner.

\cutparagraphup
\paragraph{End-to-end driving metrics (closed-loop):}
We measure the \textit{collision rate (CR)} to reflect the driving safety. This is the percentage of simulations in which there is at least 1 collision between the SDV and another actor.
We also evaluate the \textit{progress} made by the SDV on its desired route throughout the simulation horizon, measured in meters from the starting location, as well as the progress per collision, giving an idea of the ratio between non-conservativeness and safety.
Finally, we measure the mean jerk and acceleration applied as a metric of the driving comfort.
As the autonomy unrolls its own actions for long time periods, potentially diverging from the path the expert-driver executed, these metrics capture the quality of the end-to-end system, including its robustness to distributional shift \cite{ross2011reduction}.

\cutparagraphup
\paragraph{Sub-system level metrics (open-loop):}
In the open-loop evaluations, our model is evaluated on real data from the logs in the \ourdataset{} dataset (i.e., the scenes visited by the expert driver), as opposed to closed-loop evaluations where we unroll our own plans.
To evaluate the object detection quality we measure the standard \textit{mean-average precision (mAP)}, but defer the results to the supplementary because all the models share the same perception backbone architecture, and it is not the focus of this paper.
To measure the reconstruction capability and the diversity of the scene-level motion forecasts, we use $K=15$ scene samples, meaning that there are 15 distinct future scenarios predicted, each with 1 trajectory per actor.
The \textit{minimum scene average displacement error (minSADE)} measures how well we recall the ground-truth trajectory, while the \textit{mean scene average displacement error (meanSADE)} measures the precision of the predicted distribution as proposed in \cite{casas2020implicit}.
To evaluate how the diversity of these predictions impact the subsequent contingent plans, we measure the pairwise plan average self-distance (\textit{meanPlanASD}), i.e., the average distance between the contingent plans for 2 distinct futures.
Additionally, 
we also compute the scene average self-distance (\textit{meanSASD}), which computes the average pairwise distance among scene samples as a way to measure general diversity as proposed by \cite{yuan2019diverse,yuan2020dlow}.

\cutsubsectionup
\subsection{Comparison against state-of-the-art}

\cutparagraphup
\paragraph{Planning benchmark:} The closed-loop experiment results for motion planning are shown in Table~\ref{table:main_closed_loop}. \ourmodel{} outperforms the baselines in almost all metrics. In particular, we see a 21\% increase in progress per collision to the next best baseline for this metric, CVAE + PLT. This is a combination of having 8\% fewer collisions in addition to 12\% greater progress, showing our model is able to avoid dangerous scenarios on the road without slowing down (i.e., it provides additional safety while being less conservative).
For completeness, the results of the baselines paired with our contingency planner are available in the supplementary.

\cutparagraphup
\paragraph{Diversity tradeoffs:} The open-loop experiment results are shown in Fig.~\ref{fig:diversity_plots}. \ourmodel{} achieves the safest plans, makes the most progress, and achieves the best prediction reconstruction while being among the most diverse methods.
In the baselines, we see how more diversity often makes the plans more unsafe, diminishes the progress throughout the route, and regresses the reconstruction quality of the motion forecasts. In contrast, our method escapes this ``diversity curse".
Importantly, the predictions given to our contingency planner are very accurate and diverse at the same time. This allows our model to make cautious, safe plans without the shortcomings of too much irrelevant or excessively variant predictions.
Here, the baselines use the PLT planner, but we include the same plots with the contingency planner in the \supp.

\cutsubsectionup
\subsection{Ablation Study}
Table~\ref{table:ablation_closed_loop} shows the impact of our main contributions.

\cutparagraphup
\paragraph{Diverse sampler vs. Monte Carlo sampling:} $M_1$ samples independently from the the prior $p(Z)$. We can see that when using \ourmodel{}'s diverse sampler we achieve almost the same comfort and progress, while being able to anticipate and avoid significantly more collisions.

\cutparagraphup
\paragraph{Planning diversity energy:}
$M_2$ shows that the planning diversity energy is critical for increasing the driving safety (28\% lower collision rate). We also observe an improvement in jerk and a regression in lateral acceleration. We hypothesize that this energy term favors early and preventive lateral displacements instead of late hard brakes from the planner. Further investigation is left for future work.

\cutparagraphup
\paragraph{Scenario scoring vs. uniform probabilities:} $M_3$ removes the scenario scoring, assigning each diverse scenario an equal probability as input to the planner. We can see that scenario scoring improves safety and progress, showing us that it prevents the SDV from unnecessary premature braking to avoid low-probability risks.

\cutparagraphup
\paragraph{Contingency planner vs. PLT:} The ablation $M_4$ demonstrates the importance of the contingency planner as it improves almost every metric when compared to the PLT planner, notably reducing collisions by 38\%.

\cutsubsectionup
\subsection{Qualitative results} 
Figure \ref{fig:qualitative} shows three challenging scenarios the SDV encountered while driving in closed-loop simulation. We can see in each scenario that the SDV plans multiple contingent trajectories that each respond safely to one of the predicted futures. Thus, the SDV can take a non-conservative immediate action and still find a safe future trajectory if any of the on-coming or turning cars block its path.
\cutsectionup
\section{Conclusion}
\cutsectiondown
We have proposed a prediction and planning model that generates more diverse motion forecasts and safer trajectories for the SDV. Our prediction model learns to generate multimodal trajectory samples from a joint distribution over actor trajectories. Unlike previous diverse forecasting approaches, we directly optimize for predicting rare behavior that could impact the SDV, and estimate the probability distribution over these samples for more accurate risk assessment. Our contingency planner improves the decision making over these diverse samples. Our experiments on closed-loop simulations and a large-scale dataset demonstrate that our model drives safer and less conservatively than previous state-of-the-art models.

{\small
\bibliographystyle{ieee_fullname}
\bibliography{egbib}
}

\newpage

\appendix
{\noindent \Large \textbf{\Supp} \vspace{0.15cm}} \\

In  this \supp, we first describe additional implementation details, then we discuss additional evaluation details and results, and finally showcase additional qualitative results.
The supplementary video contains a narrated overview of the method and longer duration rollouts of our model driving in closed-loop simulation.

\section{Implementation Details}
In this section, we cover implementation details about the submodules of our end-to-end driving model as well as training.

\subsection{Joint Perception and Motion Forecasting Details}
Here, we discuss the implementation details for our scene-consistent join perception and motion forecasting model from sensor data.

\paragraph{Data Input Parameterization:}

The preprocessing of our LiDAR point cloud input to the model follows in the same manner as \cite{casas2018intentnet}. Primarily, we use a Bird's Eye View (BEV) of a voxelized 3D LiDAR point cloud, with the height and time dimensions being raveled into the channel dimension. Our model does not use tracks as input, so motion information is accounted for by including past LiDAR sweeps into the input. We project past LiDAR sweeps into the coordinate frame of the SDV's current LiDAR sweep, and concatenate them in the channel dimension.
Additionally, our high-definition maps are represented as a stack of rasterized images, as described in \cite{casas2018intentnet}. These maps encode elements of the road such as intersections, lanes and roads, with different elements encoded in different channels.

\paragraph{Shared Perception Backbone:}

To extract features for object detection and motion forecasting, we use a backbone network as described in \cite{casas2020implicit}, which was adapted from \cite{yang2018pixor}. We separately process the LiDAR and HD maps in two separate sets of convolutional layers, concatenate those intermediate features channel-wise (as they use the same spatial resolution and coordinate system), and fuse them with a convolutional header to get a scene-level feature map. In particular, the LiDAR backbone is composed of 4 residual convolutional blocks with 2, 2, 3, and 6 layers respectively. These blocks use 32, 64, 128, and 256 filters and a stride of 1, 2, 2, and 2 respectively. The HD map backbone is also composed of 4 residual blocks with 2, 2, 3, and 3 layers respectively. The HD map backbone uses 16, 32, 64, and 128 filters and a stride of 1, 2, 2, and 2 respectively. For both backbones, the output of each residual block is concatenated to create a final multi-resolution feature map, as detailed in \cite{yang2018pixor}. These features maps are down-sampled 4x relative to the input. Finally, we use a header network with 4 convolutional layers and 256 filters per layer to fuse the concatenated features. GroupNorm \cite{wu2018group} is used because of our small batch size (number of frames) per GPU, which we adopt due to GPU memory constraints. The final feature map is used for the detection and motion forecasting networks. Note that the backbone network and object detector architecture is shared across all models including the baselines to make the comparison more fair and direct.

\paragraph{Object Detection Header:}

To detect the actors in the scene, we input the feature map from the backbone into one convolution layer to predict a confidence score and another convolution layer to predict a bounding box for each anchor location, following the parameterization described in \cite{yang2018pixor}. Next, non-maximal suppression (NMS) with an IoU of 0.1 is used to filter overlapping detections and low probability detections are filtered by a threshold corresponding to the maximum F1 score across the Precision-Recall curve, to arrive at a final set of actor bounding boxes.

\paragraph{Actor Feature Extraction:}

Finally, in order to obtain local actor contexts $x_n^{local}$, we use rotated ROI Align \cite{huang2018improving} and extract a crop of the feature map of a fixed size around each detected actor, which is rotated to align with the actor's centroid and orientation. The cropped region spans 10m to the back, 70m to the front, and 40m to each side of the actor, and has dimension 40 x 40 x 256. We apply an \textit{actor CNN} (a 4-layer  convolutional network with heavy downsampling) to each feature map to get a 512-dimensional feature vector $x_n^{local}$ for every actor. In order to incorporate global information about the actor's pose in the context of the whole scene, we add the BEV centroid and rotation relative to the SDV $x_n^{global} = \{c_{x,n}, c_{y, n}, a_n\}$ as features, and concatenate these two feature vectors channel wise to get the final actor context $x_n = [x_n^{local}, x_n^{global}] \in \mathbb{R}^D$

\paragraph{Scene Interaction Module:}

For the basis of our motion forecasting model, we use a "scene interaction module" (SIM) graph neural network as described in \cite{casas2020implicit} and inspired by \cite{casas2019spatially}. SIM is used in the Prior, Encoder, Decoder, Diverse Sampling networks as well as the Scenario scorer. Given a graph of actor nodes with embeddings, the SIM will pass in the hidden states of the two actor nodes in a given edge, in addition to the projected distance between their two bounding boxes, to a 3-layer MLP. This computes an activation for each edge in the graph, that goes through feature-wise max-pooling and a GRU cell to compute a new hidden state for each node. Finally, a 2-layer MLP is applied to each node to get their final outputs.
All our SIMs use a hidden state size of 64. We run two SIMs in parallel for each of our Prior and Encoder networks to compute a latent $\mu$ and $\sigma$ vector of length 64 for each actor. These are sampled in a gaussian parameterization to get $z_n$ vectors of length 64 for each agent. These $z_n$ vectors and the $x_n$ feature vector is then concatenated for each actor for a total length of 576. This set of actor vectors is then fed into a decoder SIM that computes a length 20 output ($(x, y)$ waypoints over 10 time steps).

\subsection{Planning-Centric Diverse Sampler Details}
Here, we will describe the implementation details of our diverse sampler network.

\paragraph{Network Architecture:}

To train the diverse sampler network $\mathcal{M}_\eta$, we first train the scene-consistent joint perception and motion forecasting model described above, and freeze the detection backbone, actor feature extractors and decoder network. We replace the Encoder and Prior networks with our diverse sampler. This model consists of two SIMs, one for the $A$ vector, and another for the $B$ vector. These SIMs have identical architecture - they take as input the graph of actor feature vectors of length 512, and use hidden states of length 64. Each model jointly outputs $S=15$ samples of 64-dimensional vectors for each agent, and does this by outputting a $S * 64$ length vector for each agent. As described in section 3.2, these vectors parameterize the latent mapping that is used to sample $\mathbf{Z}$, a set of $S$ latent vectors. These are then decoded into $S$ scene predictions, which is done in parallel by batching the $S$ latent samples as input to the decoder.

As mentioned in the description of the DLow baseline, we stray from the implementation described in \cite{yuan2020dlow} by predicting only the diagonals of the $A$ matrix instead of a full matrix, because the full matrix would not fit in memory given our high dimensionality. We have higher dimensionality because our scene latent vectors represent a sample of the joint distribution of all actors, as opposed to the marginal distribution of a single actor.

\paragraph{Hyperparameters:}
For learning, we weight the energies in our model as follows: $E_r$ has a coefficient of $0.02$, $E_p$ has a coefficient of $0.01$, $E_d$ has a coefficient of $10$, and $\sigma_d$ is equal to $10000$.

\subsection{Scenario Probability Estimation Details}
The scenario probability estimation network is parameterized as another SIM with a hidden dimension of 128. 
This SIM takes as input the $S$ predicted scenarios $\mathbf{Y}$.
To do so, the $n$-th node state in the graph is initialized to the $S$ trajectories of actor $n$, $\mathbf{Y}_n \in \mathbb{R}^{2TS}$. After 1 round of message passing in the SIM, we take all the updated node states and average pool them over the node (or actor) dimension, thus obtaining a single feature vector of the hidden dimension (128).
Then, a MLP maps these features into the $S$ scores, one for each future.
\subsection{Contingency Planner Details}
In this section we provide details of the action and trajectory sampling, followed by the description of the planner cost functions.

\subsubsection{Action and trajectory sampling}
Since the motion-paths representing the lane centers are strong priors for potential SDV paths, we perform the action and trajectory sampling
in Frenet Frame of the goal motion-path, given by the input route. The action and its corresponding set of long-term trajectories are represented by
lateral and longitudinal trajectories relative to the goal motion-path \cite{werling2010optimal}. The sampling is achieved by first generating a lateral profile. We use
two quintic polynomials that are generated by the initial SDV state in Frenet frame, and sampled lateral offsets for mid/end-conditions as in \cite{sadat2019jointly}.
Next, to generate the actions, we sample longitudinal profiles in form of quartic polynomials which, combined with the generated lateral trajectory, yields
a bicycle model trajectory representation. Similarly, in order to sample contingent plans, we generate long-term longitudinal trajectories in form of two quartic
polynomials. These polynomials are conditioned on the end longitudinal-state of the corresponding action, and sampled mid/end-conditions \cite{sadat2019jointly}.
Note that we use 1 second horizon for the actions and 4 seconds for the trajectories.

\subsubsection{Costing}
The planner cost function includes subcosts that encode different aspects of the sample actions and trajectories, including safety, traffic-rules, and comfort.

\cutparagraphup
\paragraph{Safety:}
Given the predicted trajectories of the actors, the collision subcost penalizes a sample SDV trajectory if the SDV polygon is overlapping with the polygon of the other actors. This collision cost is computed separately for each class of actors. Furthermore, a trajectory is penalized if it has high velocity close to other actors.
Another subcost related to safety is the headway subcost, in which the SDV trajectory is penalized if it is violating a safety distance to the leading vehicle. This safety distance is determined by the velocity of both SDV and the lead vehicle such that the SDV can stop with a comfortable acceleration profile, in case the lead vehicle suddenly stops with hard breaking.

\cutparagraphup
\paragraph{Traffic rules:}
The SDV is required to stay on its lane and close to the centerline. Therefore, trajectories that are far from the lane-motion paths are penalized proportional to the offset. Similarly, if the SDV polygon goes off of the lane, the trajectory is penalized. In order to prevent the SDV from violating red-lights, trajectories that enter red-light intersections are penalized proportional to the violation distance. We use similar costing for junctions with stop-signs. Furthermore, trajectories that go above the speed-limit of the road are penalized proportional to the violation margin.

\cutparagraphup
\paragraph{Progress and comfort:}
In order to promote trajectory samples that progress in the route, we use the
traveled longitudinal distance as a reward (negative cost).
Additionally, trajectories that violate the kinematic and dynamic constraints of the vehicle are penalized, including curvature, acceleration, deceleration, and lateral acceleration. Additionally, high jerk and acceleration and decelerations are penalized by cost functions to promote comfortable trajectories.

\subsection{Optimization Details}
When training the scene-consistent motion forecaster, we used the same optimization settings as in \cite{casas2020implicit}, where we use the Adam optimizer \cite{kingma2014adam} with a learning rate of 1.25e-5, with a cyclical annealing schedule for one of our coefficients. This training ran for 50,000 iterations of batch size 4 on 16 Nvidia RTX 5000 GPUs. When training the diverse sampler, we use the same learning rate, without cyclical annealing schedule, training this model for 40,000 iterations of batch size 1 (due to memory constraints) on 8 Nvidia RTX 5000 GPUs.
\section{Additional Evaluation Details}
\subsection{Operating point for evaluation}
We evaluate motion forecasting on true positive detections. To find a fair operating point of the object detectors for all models in the motion forecasting task, we follow \cite{casas2019spatially,casas2020implicit} and find the detection threshold corresponding to a common recall point. In particular, we evaluate motion forecasting at $90\%$ recall for vehicles, $60\%$ for bicyclists and $70\%$ for pedestrians.

For the downstream task evaluation of motion planning, we find the maximum F1 score point in the Precision-Recall curve for each baseline, and operate the detector at that point to minimize false positive and false negatives.

\cutsubsectionup
\subsection{Formal sub-system level metrics definitions}

The metrics used at the sub-system level in our open-loop evaluation are defined below.
The \textit{minimum scene average displacement error (minSADE)} measures how well we recall the ground-truth trajectory by measuring the distance between the ground-truth scene and the closest predicted scene.
The \textit{mean scene average displacement error (meanSADE)} measures the precision of the predicted distribution at the scene-level as proposed in \cite{casas2020implicit} by measuring how different the predicted scenes are on average with the ground-truth.

\cutequationup
{\small
\begin{align}
    \text{minSADE} &= \min_{s \in 1...S} \frac{1}{NT} \sum^N_{n=1} \sum^T_{t=1} || y^t_{n, GT} - y^t_{n, s}||_2,\\
    \text{meanSADE} &=  \frac{1}{NTS} \sum^S_{s=1}\sum^N_{n=1} \sum^T_{t=1} || y^t_{n, GT} - y^t_{n, s}||_2,
    \end{align}
}
where $N$ is the number of actors, $T$ is the number of timesteps, $S$ is the number of scene samples for a given scenario, $y_{n, s}$ is the predicted trajectory for actor $n$ in the scene sample $s$, $y_{n, GT}$ is the ground truth trajectory for actor $n$.

We focused on measuring motion forecasting diversity that impacts the safety of the SDV, by evaluating how the diversity of these predictions impact the subsequent contingent plans. To do this, we measure the pairwise plan average self-distance (\textit{meanPlanASD}), i.e., the average distance between the contingent plans for 2 distinct futures.
Additionally, we also compute the scene average self-distance (\textit{meanSASD}), which computes the average pairwise distance among scene samples, and the minimum scene self-distance (\textit{minSASD}), which computes the minimum pairwise distance for each scene sample, as ways to measure general diversity as proposed by \cite{yuan2019diverse,yuan2020dlow}.

\cutequationup
{\small
\begin{align}
    \text{meanPlanASD} &= \frac{1}{S} \sum^S_{i=1}\sum^S_{j \neq i} \ell_2(\tau_i, \tau_j)
\end{align}
}
where $\tau_i = \tau(Y_i)$ is the corresponding planned contingent trajectory to the scene-level future $Y_i$.

\cutequationup
{\small
\begin{align}
    \text{meanSASD} &= \frac{1}{S} \sum^S_{i=1} \sum^S_{j \neq i} \ell_2(Y_i, Y_j) \\
    \text{minSASD} &= \sum^S_{i=1} \min_{s \in 1...S} \ell_2(Y_i, Y_s)
\end{align}
}
where $Y_i$ is the tensor of coordinates of the future trajectories in scene $i$, with dimensions $(N, S, T, 2)$.

\cutsubsectionup
\subsection{Baselines}

All of these baselines share the shared perception backbone, object detection, and actor feature extraction models as described above. These baselines are divided into actor-independent models that either output explicit marginal likelihoods (i.e. MultiPath \cite{chai2019multipath}) or output sampled trajectories such as our CVAE\cite{sohn2015learning}, DPP \cite{yuan2019diverse} and DLow \cite{yuan2020dlow} models, and scene-consistent models that are autoregressive (ESP \cite{rhinehart2018r2p2}) or implicit variable models (ILVM \cite{casas2020implicit}).

As detailed in \cite{casas2020implicit}, for MultiPath, we use their mixture of trajectories parameterization instead of our encoder-decoder architecture, to predict a gaussian for each waypoint.

For the CVAE model, we replace the Encoder, Prior and Decoder SIMs with MLPs of similar dimension, but use the same variational inference parameterization.

For the DPP model, we use the frozen Decoder from the CVAE model. Similarly to \cite{yuan2019diverse}, instead of predicting a $\mu$ and $\sigma$ vectors, we predict $S$ scenes samples of latent vectors using an MLP, where each scene sample consists of a 64-dimensional latent vector for each actor. We do this by predicting a $S * 64$-dimensional vector for each actor, and reshaping that into $S$ vectors of length $64$ per actor. Then, we decode each scene sample of latent vectors to get $S$ separate scene motion forecasts $Y$. We then apply the determinantal point process loss as described in \cite{yuan2019diverse}. To get the $x_i$ vectors, as referred to their work (not to be confused with the \ourmodel{} definition), we concatenate the trajectories in one scene over all the actors, their $(x, y)$ coordinates over time for each scene to get a $N * T * 2$-dimensional vector $x_i$ for each scene $i$, where $N$ is the number of actors and $T$ is the number of future timesteps predicted. To get the $z_i$ vectors, as referred to their work (not to be confused with the \ourmodel{} definition), we concatenate the per-actor vectors in each scene to get $N * 64$-dimensional vectors $z_i$ for each scene $i$. We use these vectors to compute the Diversity Loss as described in their paper.

For the DLow model, we use the frozen Decoder from the CVAE model. Similarly to \cite{yuan2020dlow}, we predict $S$ scene samples of an $A$ and $B$ vector for each agent using an MLP. Unlike their paper, we output an $A$ vector representing the diagonal of the matrix, because the full matrix would not fit in memory given our high dimensionality. These $A$ and $B$ are essentially used in the same way as described in section 3.2 of our paper, except planning-based diversity and scenario probability scoring are not used.

For our ESP model, we adapt it to our feature contexts and memory constraints as described in \cite{casas2020implicit}.

For our ILVM baseline, we use exactly the formulation described in \cite{casas2020implicit} (where SIMs are used for the Encoder, Prior and Decoder), except we use a fixed standard gaussian distribution as our target encoder distribution instead leaving it unconstrained. The same is true for our CVAE model, and \ourmodel{}. This is because it allows our DPP, DLow and \ourmodel{} models to more easily learn to fit and map to the distribution of the latent space, making training much more tractable.

\section{Additional Evaluation Results}

\newcolumntype{s}{>{\centering\arraybackslash}X}
\begin{table*}[t]
	\centering
	\begin{threeparttable}
        \begin{tabularx}{\textwidth}{
                        l |
                        l |
                        s s s s s s s s%
                        }
		    \toprule
              \textbf{Prediction} & \textbf{Planning} &
              CR$(\%)$  &
              $\frac{\text{Progress}}{collision}(m)$&
              Progress$(m)$&
              Jerk$(\frac{m}{s^3})$
              & Lat.Acc.$(\frac{m}{s^2})$
              & Acc$(\frac{m}{s^2})$
              & Decel$(\frac{m}{s^2})$  \\
            \midrule
\multirow{2}{*}{CVAE-DPP} & PLT & 17.07 & 123.97 & 21.17 & 11.99 & \textbf{0.06} & 1.11 & 0.80 \\
 & Contingency & 12.80 & 235.21 & 30.12 & 8.83 & 0.16 & 1.03 & 0.54 \\
 \midrule
\multirow{2}{*}{CVAE-DLow} & PLT & 14.63 & 377.07 & 55.18 & 5.22 & 0.15 & 0.84 & 0.54 \\
 & Contingency & 9.76 & 628.67 & 61.33 & 4.44 & 0.36 & 0.79 & 0.36 \\
 \midrule
\multirow{2}{*}{MultiPath} & PLT & 12.20 & 394.37 & 48.09 & 12.92 & 0.13 & 1.24 & 0.80 \\
 & Contingency & 10.37 & 548.72 & 56.88 & 7.62 & 0.33 & 1.08 & 0.57 \\
 \midrule
\multirow{2}{*}{ESP} & PLT & 11.59 & 464.44 & 53.81 & 6.52 & 0.15 & 0.89 & 0.57 \\
 & Contingency & 10.98 & 549.89 & 60.35 & 5.20 & 0.35 & 0.82 & 0.53 \\
 \midrule
\multirow{2}{*}{ILVM} & PLT & 10.98 & 553.96 & 60.80 & 5.50 & 0.16 & 0.86 & 0.56 \\
 & Contingency & 9.15 & 709.60 & \textbf{64.90} & \textbf{4.40} & 0.38 & \textbf{0.77} & \textbf{0.52} \\
 \midrule
\multirow{2}{*}{CVAE} & PLT & 8.54 & 655.22 & 55.93 & 7.22 & 0.15 & 0.96 & 0.62 \\
 & Contingency & 9.76 & 630.50 & 61.51 & 5.07 & 0.36 & 0.82 & 0.53 \\
\midrule
\multicolumn{2}{l|}{LookOut} & \textbf{7.93} & \textbf{790.37} & 62.65 & 4.69 & 0.37 & 0.79 & 0.53 \\
	  \bottomrule
		\end{tabularx}
    \end{threeparttable}
    \caption{\textbf{Closed loop motion planning comparison against the baselines with Contingency Planner.} }
	\label{table:main_closed_loop_including_contigency}
\end{table*}

\paragraph{Closed-loop experiments with baselines with Contingency Planner:}
We see in Table~\ref{table:main_closed_loop_including_contigency} that the contingency planner increases the progress and decreases the acceleration and deceleration of the motion planning for all baselines. Additionally, it increases the lateral acceleration for all baselines, possibly in order to make more active maneuvers to nudge around obstacles. We see that the \ourmodel{} still maintains the best safety and progress per collision even when the baselines are paired with our contingency planner, and has similar values in other metrics to the other most competitive baseline, ILVM + Contingency Planner. This results demonstrates the necessicity of using both scene-consistent diverse predictions, and the contingency planner.

\newcolumntype{s}{>{\centering\arraybackslash}X}
\begin{table*}[t]
	\centering
	\begin{threeparttable}
        \begin{tabularx}{\textwidth}{
                    l |  %
                    l |  %
                    s s s s %
                    }
		    \toprule
            Category & Model & minSADE (m) & meanSADE (m) & minSASD (m) & meanSASD (m) \\ %
            \midrule
                \multirow{7}{1.5cm}{Vehicles}
                &MultiPath  & 0.929       & 1.314        & 1.175       & 4.628    \\ %
                &CVAE       & 0.804       & 1.083        & 0.488       & 2.693    \\ %
                &CVAE-DPP   & 1.143       & 4.267        & 3.551      & 19.849    \\ %
                &CVAE-DLow  & 0.839       & 1.152        & 0.559       & 3.277    \\ %
                &ESP        & 1.090        & 1.441         & 1.120        & 3.991   \\ %
                &ILVM       & 0.770        & 1.061         & 0.455        & 2.534   \\ %
                & \ourmodel{} & \textbf{0.765}  & \textbf{1.022} & 0.704 & 4.078 \\ %
            \midrule
                \multirow{7}{1.5cm}{Pedestrians}
                &MultiPath & 0.531 & 0.691 & 0.619 & 1.960 \\
                &CVAE & \textbf{0.527} & \textbf{0.550} & 0.075 & 0.284 \\
                &CVAE-DPP & 0.668 & 3.748 & 4.477 & 17.511 \\
                &CVAE-DLow & 0.552 & 0.556 & 0.018 & 0.102 \\
                &ESP & 0.547 & 0.637 & 0.750 & 1.424 \\
                &ILVM & 0.562 & 0.576 & 0.069 & 0.239 \\
                & \ourmodel{} & 0.583 & 0.582 & 0.085 & 0.811 \\
            \midrule
                \multirow{7}{1.5cm}{Bicyclists}
                &MultiPath & 0.480 & 0.708 & 0.675 & 2.244 \\
                &CVAE & 0.514 & 0.636 & 0.249 & 0.963 \\
                &CVAE-DPP & 0.553 & 3.890 & 4.476 & 17.797 \\
                &CVAE-DLow & 0.510 & 0.629 & 0.103 & 0.574 \\
                &ESP & 0.601 & 0.925 & 0.770 & 2.239 \\
                &ILVM & \textbf{0.465} & 0.633 & 0.178 & 0.807 \\
                & \ourmodel{} & 0.510 & \textbf{0.627} & 0.164 & 0.770 \\
	        \bottomrule
		\end{tabularx}
    \end{threeparttable}
    \caption{\textbf{Multi-class motion forecasting results} in \ourdataset{} ($S=15$ samples).}
	\label{table:main_prediction_vehicle}
\end{table*}

\paragraph{Multi-class motion forecasting comparison in \ourdataset{}:}
We can see in Table~\ref{table:main_prediction_vehicle} that \ourmodel{}'s prediction module most accurately models the ground truth future trajectories for vehicles, with the lowest minimum and mean scene average displacement error. This is important because vehicles make up the vast majority of actors on the road. Compared to the baseline with the next lowest error, ILVM, we have much greater prediction diversity (61\% greater meanSASD).

We can see that for pedestrians and bicyclists, our model maintains a competitive accuracy/diversity tradeoff. We leave the problem of improving accuracy while maintaining the diversity for pedestrians and bicyclists while not regressing in vehicles for future work.

\begin{table*}[t]
    \vspace{-5pt}
    \centering
    \begin{tabularx}{\textwidth}{l | ss | ss | ss}
        \toprule
        Model &\multicolumn{2}{c|}{mAP (\%) Vehicles} &\multicolumn{2}{c|}{mAP (\%) Pedestrians} &\multicolumn{2}{c}{mAP (\%) Bicyclists} \\
        &IoU 0.5 &IoU 0.7 &IoU 0.1 &IoU 0.3 &IoU 0.1 &IoU 0.3 \\
        \toprule
        MultiPath &93.95 &82.77 &78.47 &75.48 &62.72 &56.19 \\
        ESP &95.10 &84.95 &80.97 &77.57 &70.21 &63.29 \\
        CVAE &95.76 &87.98 &\textbf{81.48} &\textbf{78.67} &74.31 &68.76 \\
        \midrule
        \ourmodel{} &\textbf{95.80} &\textbf{87.99} &80.66 &78.32 &\textbf{75.18} &\textbf{69.48} \\
        \bottomrule
    \end{tabularx}
    \caption{\textbf{Multi-class detection results} in \ourdataset{}.}
    \vspace{-5pt}
    \label{table:main_detection_table}
    \vspace{-5pt}
\end{table*}

\paragraph{Detection comparison in \ourdataset{}:} We see in Table~\ref{table:main_detection_table} that our detector's mAP on vehicles is 0.958 and is the highest among competitors for vehicles and bicyclists, and is comparable to the best performing baseline on pedestrians.
The detection performance amongst most models is similar since they all share the perception backbone network, as explained in Section 2.3. The difference stems from the differences in prediction loss in the joint perception and prediction training (stage 1 in \ourmodel{}).

\begin{figure*}[t]
    \centering
    \begin{tabular} {c@{\hspace{.5em}}c@{\hspace{.5em}}c@{\hspace{.5em}}c}
        \raisebox{-0.5\height}{\includegraphics[width=0.4\linewidth]{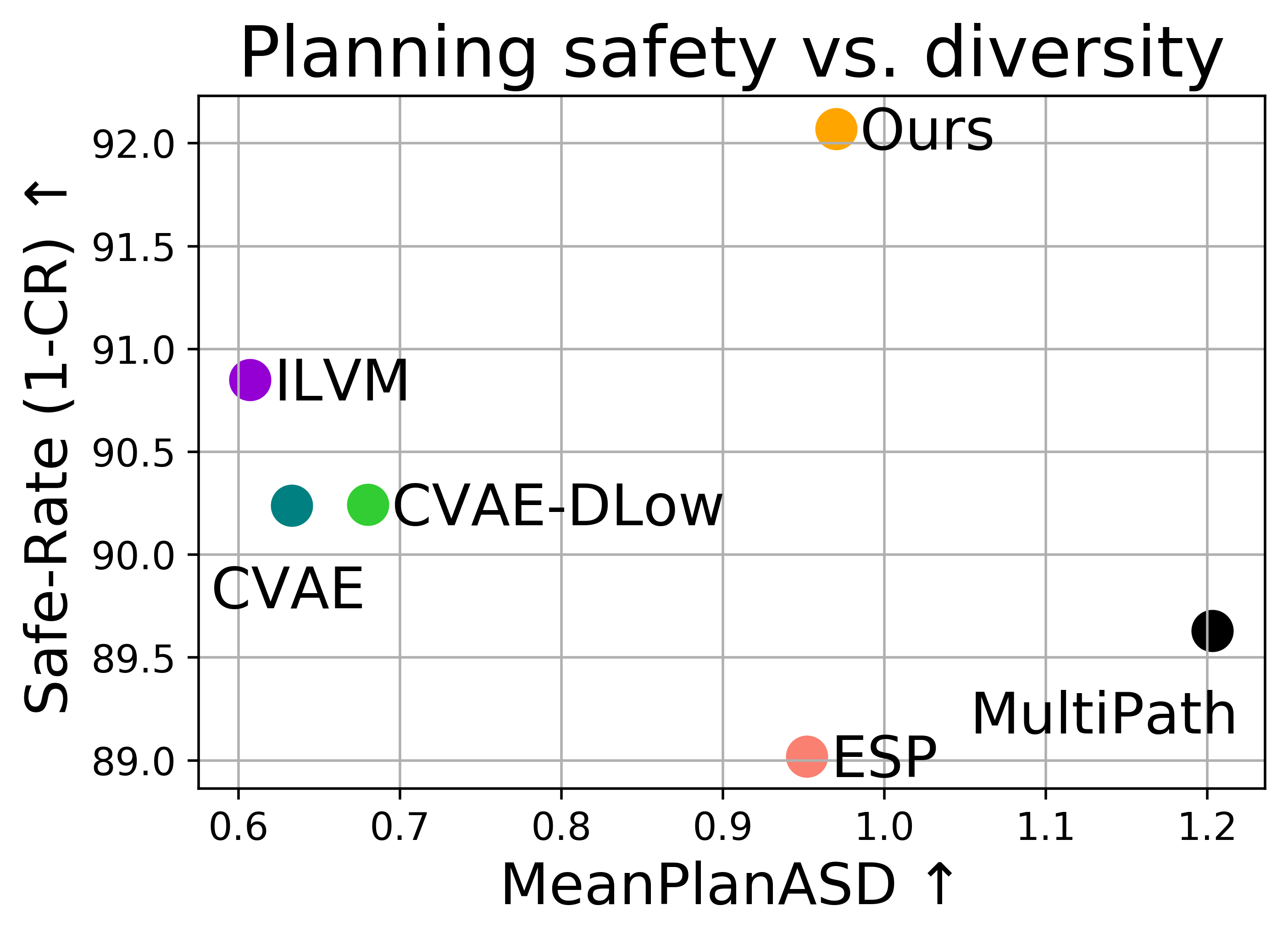}} &
        \raisebox{-0.5\height}{\includegraphics[width=0.4\linewidth]{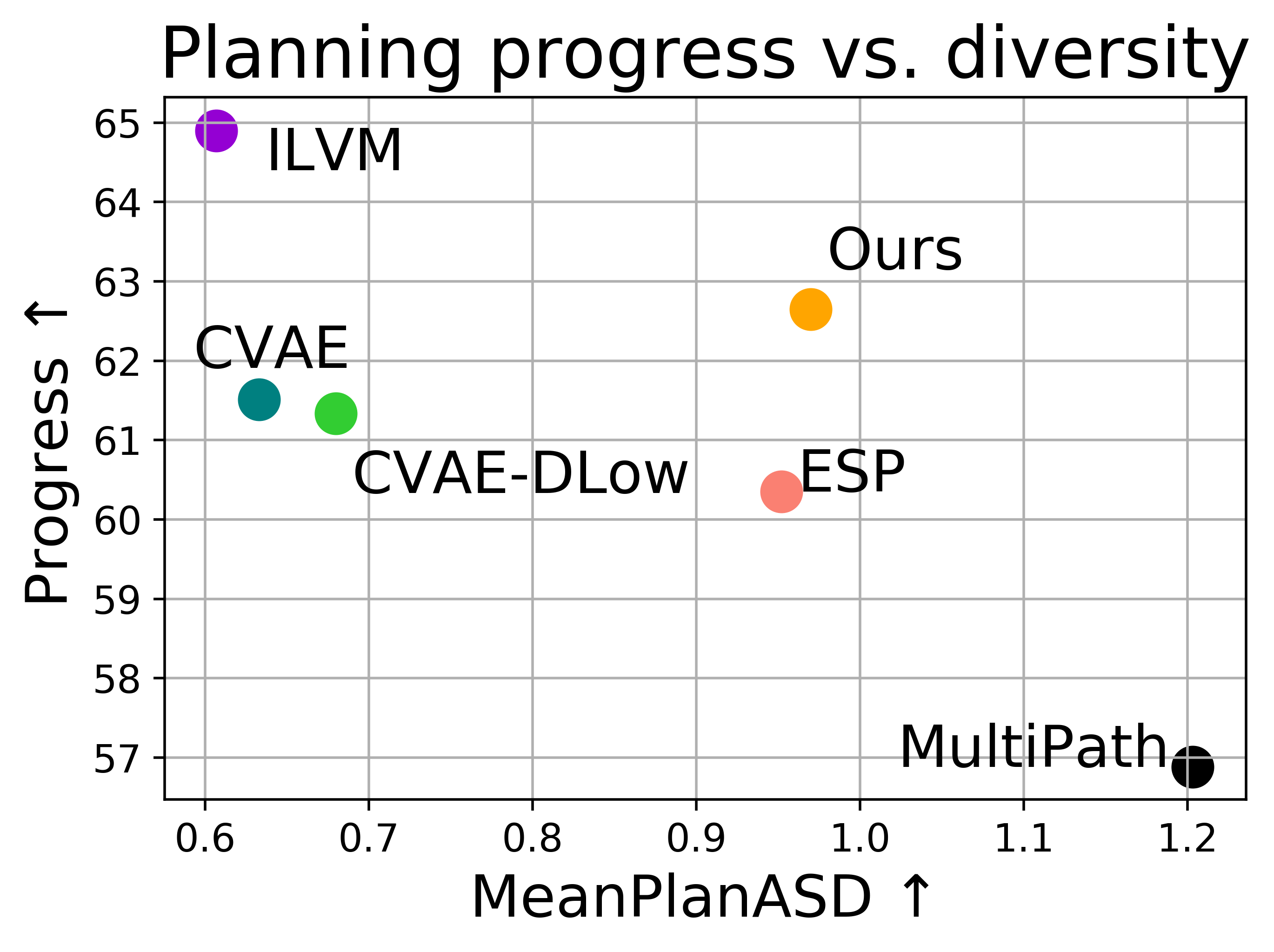}} \vspace{.5em} \\
    \end{tabular}
    \caption{
        \textbf{Planning quality as a function of diversity} when we pair the baseline prediction models with \textbf{our contingency planner}.
    }
    \label{fig:diversity_plots_contingency}
\end{figure*}
\paragraph{Planning vs. diversity tradeoffs when using contingency planner:}

We see in Fig~\ref{fig:diversity_plots_contingency} that \ourmodel{} offers the safest plans in closed-loop experiments, even when we pair the baselines with the contingency planner. We achieve a strong tradeoff between safety and meanPlanASD, with a 3\% absolute improvement in safe-rate (or equivalently, 24\% lower rate of collisions) than the only baseline that has greater planning diversity, MultiPath. This shows that our model is generating diverse predictions that are relevant to the SDV and accurately anticipate possible safety-critical scenarios the SDV should prepare for.

In addition, we see that our SDV demonstrates a competitive tradeoff between progress and planning diversity in Fig~\ref{fig:diversity_plots_contingency}. Qualitatively, we see our motion planning not make as much progress compared to the ILVM baseline, in order to slow down in situations of uncertainty to avoid collisions. On the other hand, having too many unrealistic, varied trajectories and a lack of realistic trajectories can result in a low rate of progress (such as in MultiPath) because the SDV struggles to avoid all the predicted trajectories of other actors.

\section{Additional Visualizations}
\begin{figure*}[t]
    
    \begin{tabular}{c@{\hspace{.5em}}c}
        \textbf{Scenario 1} &
        \textbf{Scenario 2} \\
        \raisebox{-0.5\height}{\includegraphics[trim={1cm .33cm 1cm .33cm},clip, width=0.45\textwidth]{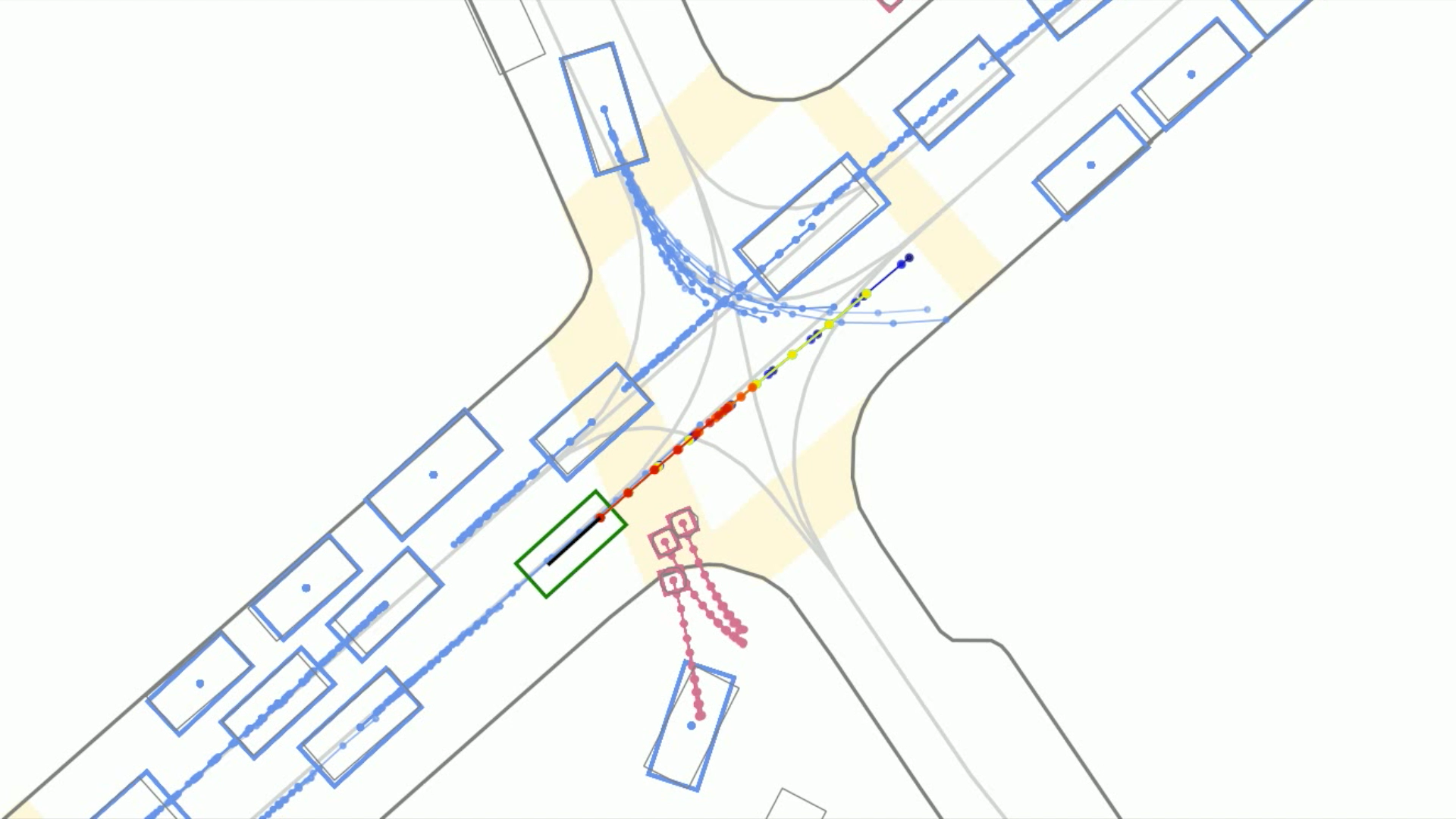}} &
        \raisebox{-0.5\height}{\includegraphics[trim={1cm .33cm 1cm .33cm},clip, width=0.45\textwidth]{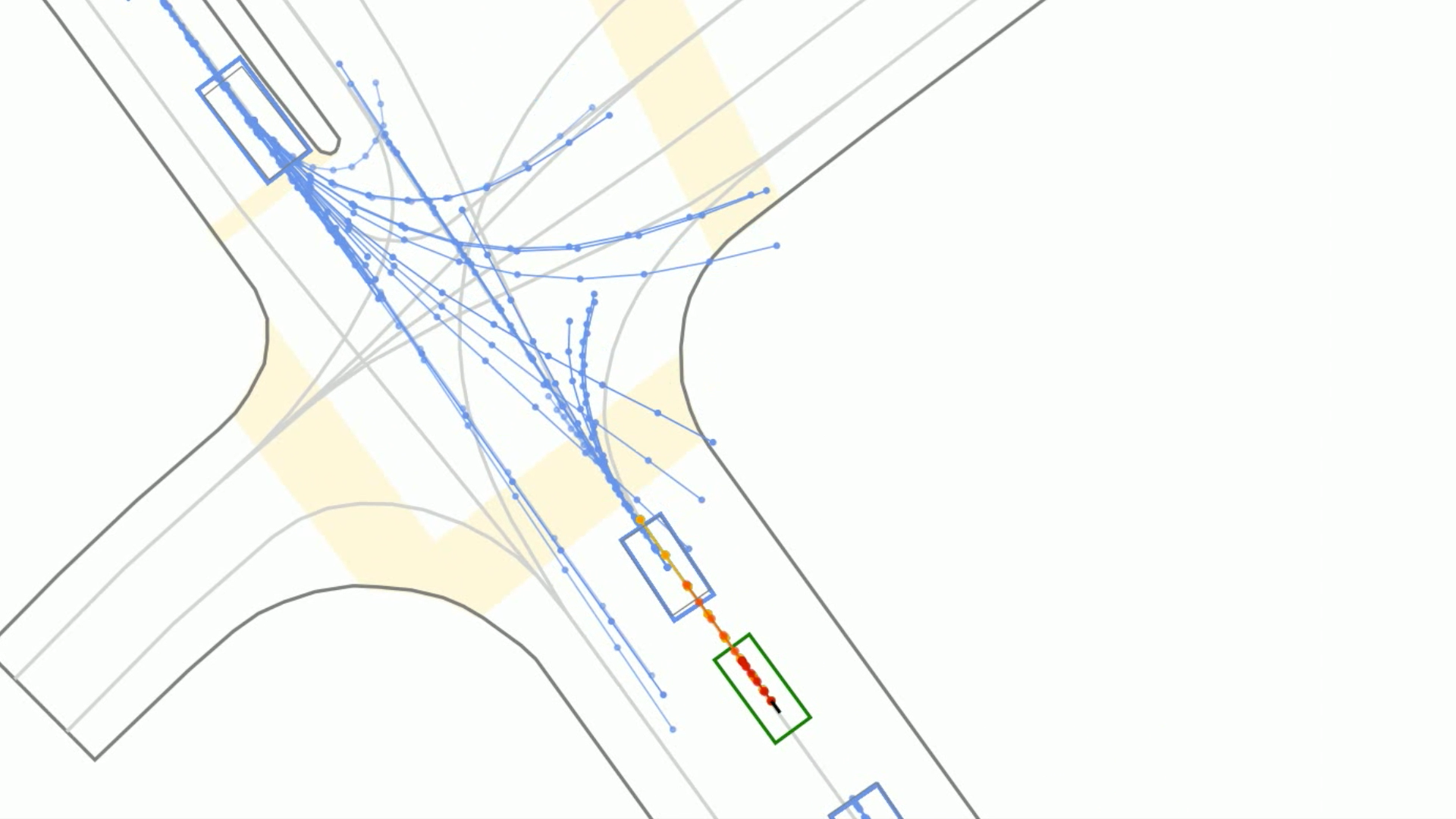}} \\
        \vspace{1cm} \\
        \textbf{Scenario 3} &
        \textbf{Scenario 4} \\
        \raisebox{-0.5\height}{\includegraphics[trim={1cm .33cm 1cm .33cm},clip, width=0.45\textwidth]{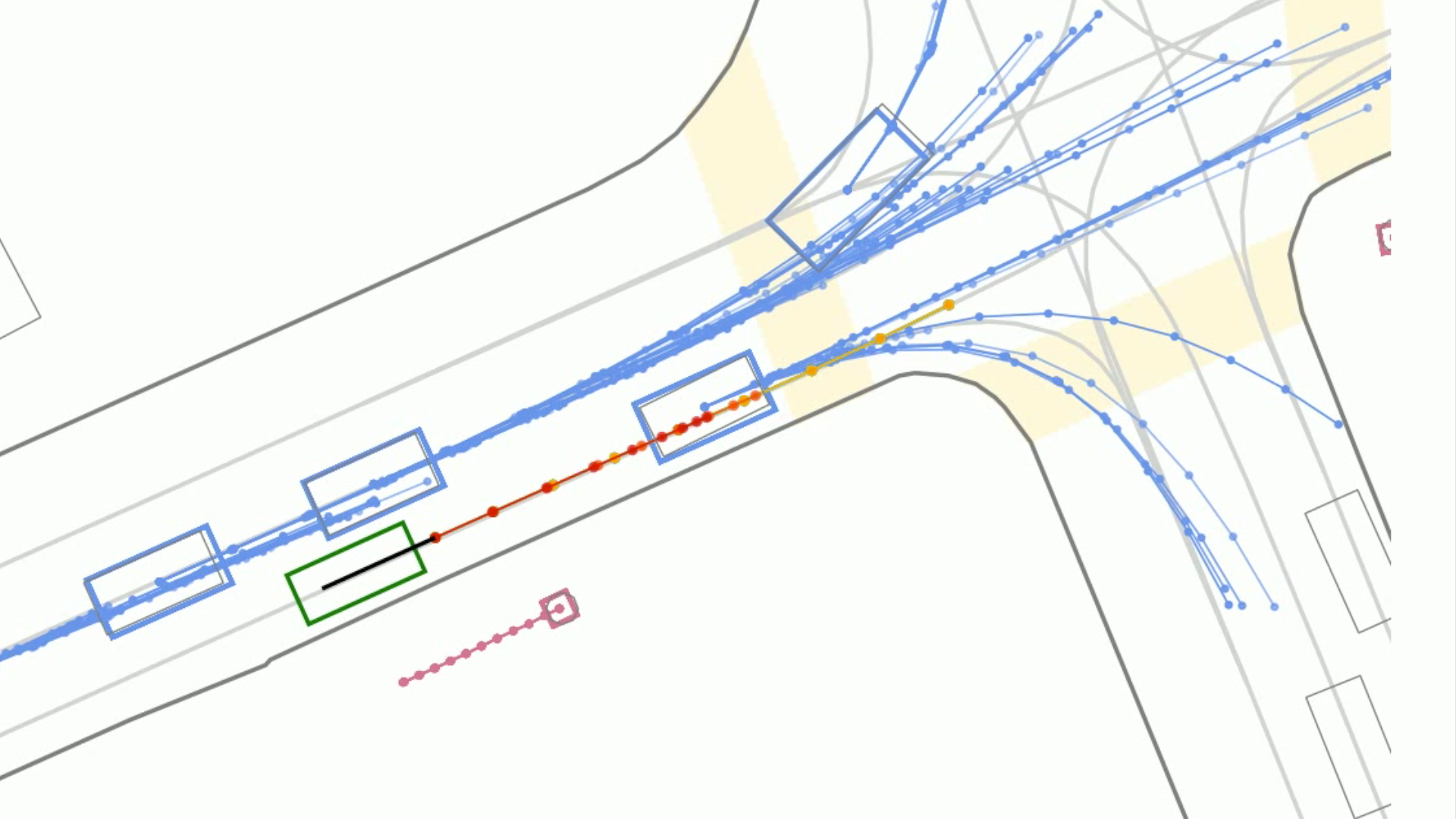}} &
        \raisebox{-0.5\height}{\includegraphics[trim={1cm .33cm 1cm .33cm},clip, width=0.45\textwidth]{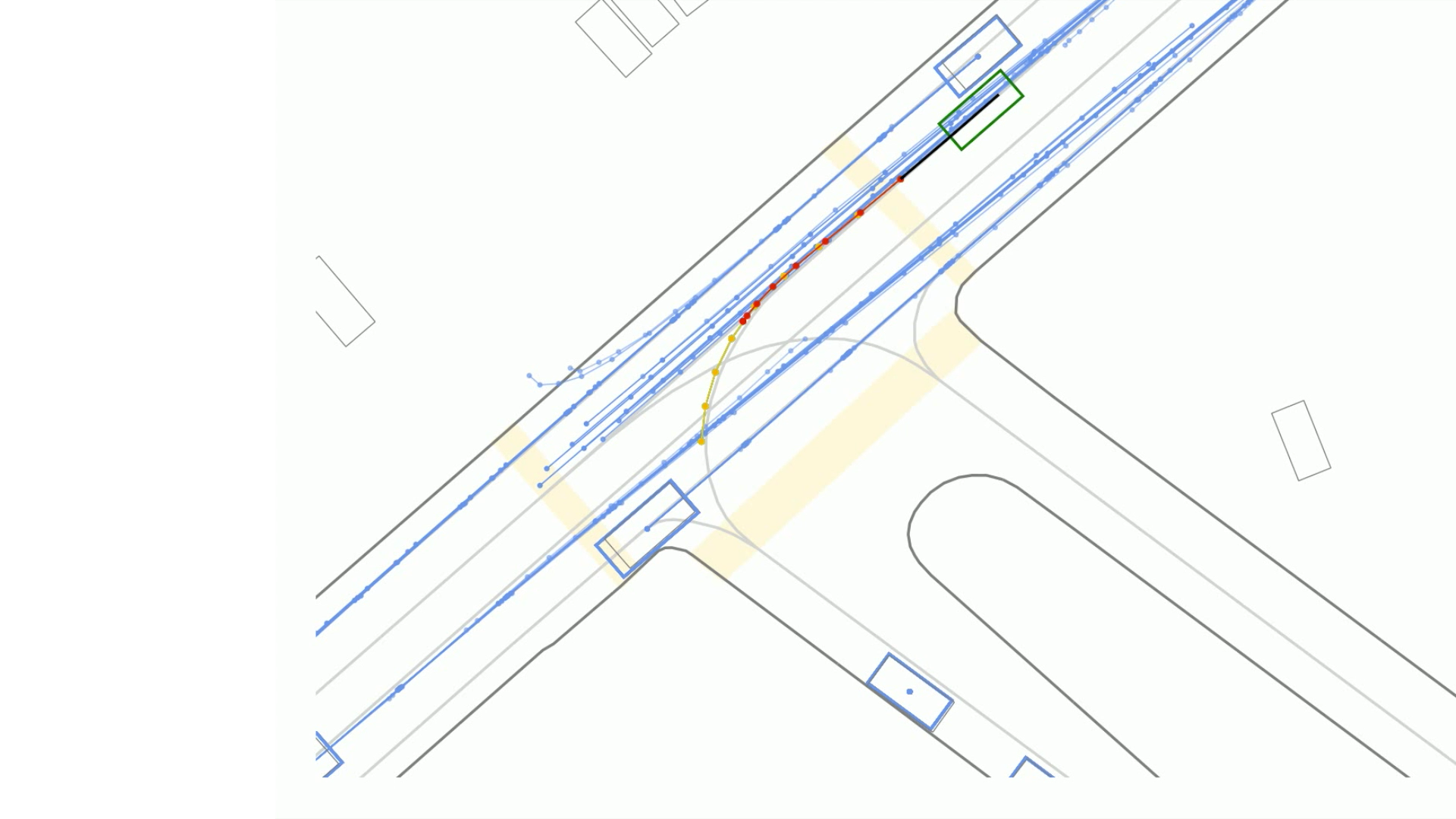}} \\
        \vspace{1cm} \\
        \textbf{Scenario 5} &
        \textbf{Scenario 6} \\
        \raisebox{-0.5\height}{\includegraphics[trim={1cm .33cm 1cm .33cm},clip, width=0.45\textwidth]{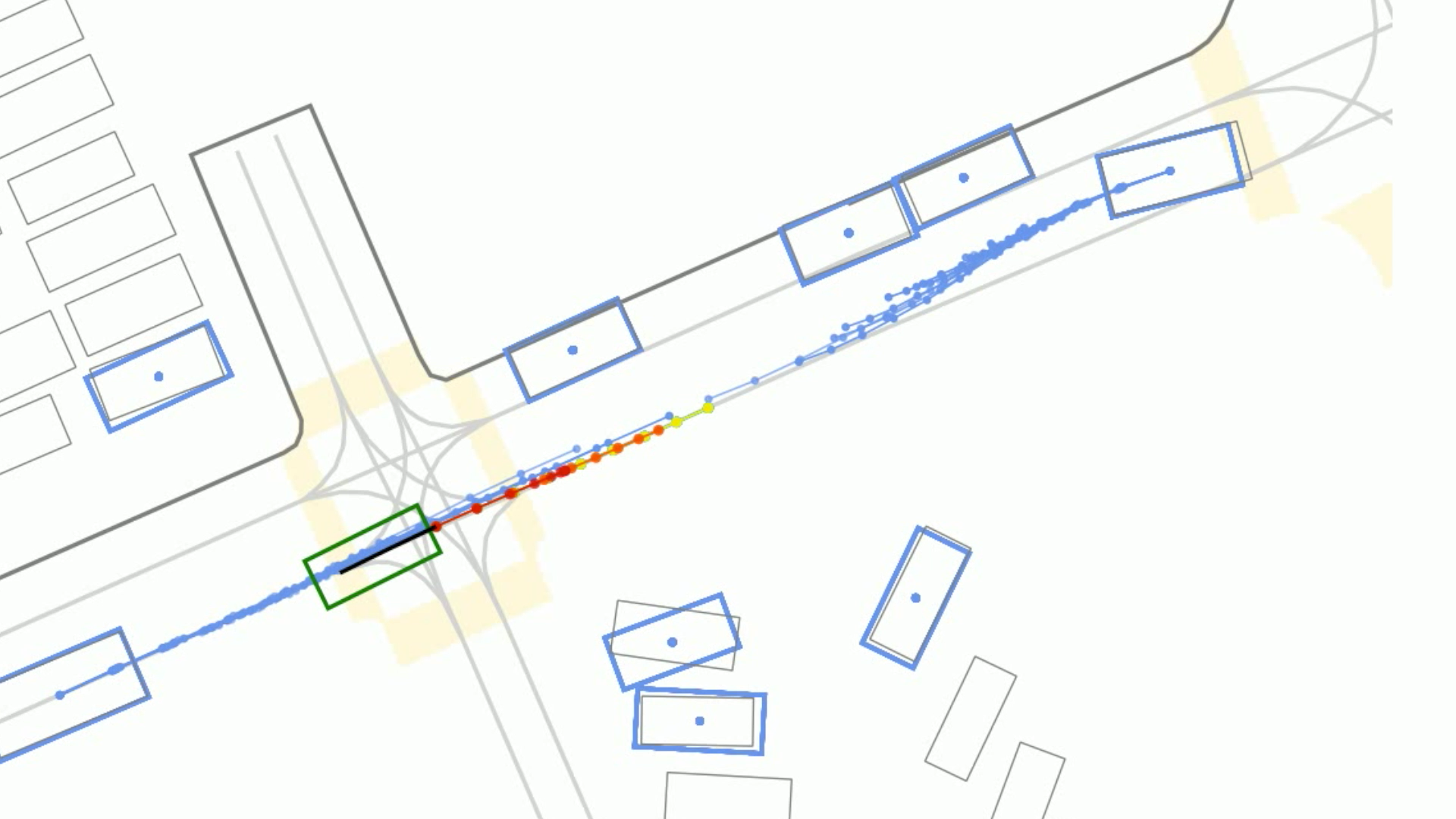}} &
        \raisebox{-0.5\height}{\includegraphics[trim={1cm .33cm 1cm .33cm},clip, width=0.45\textwidth]{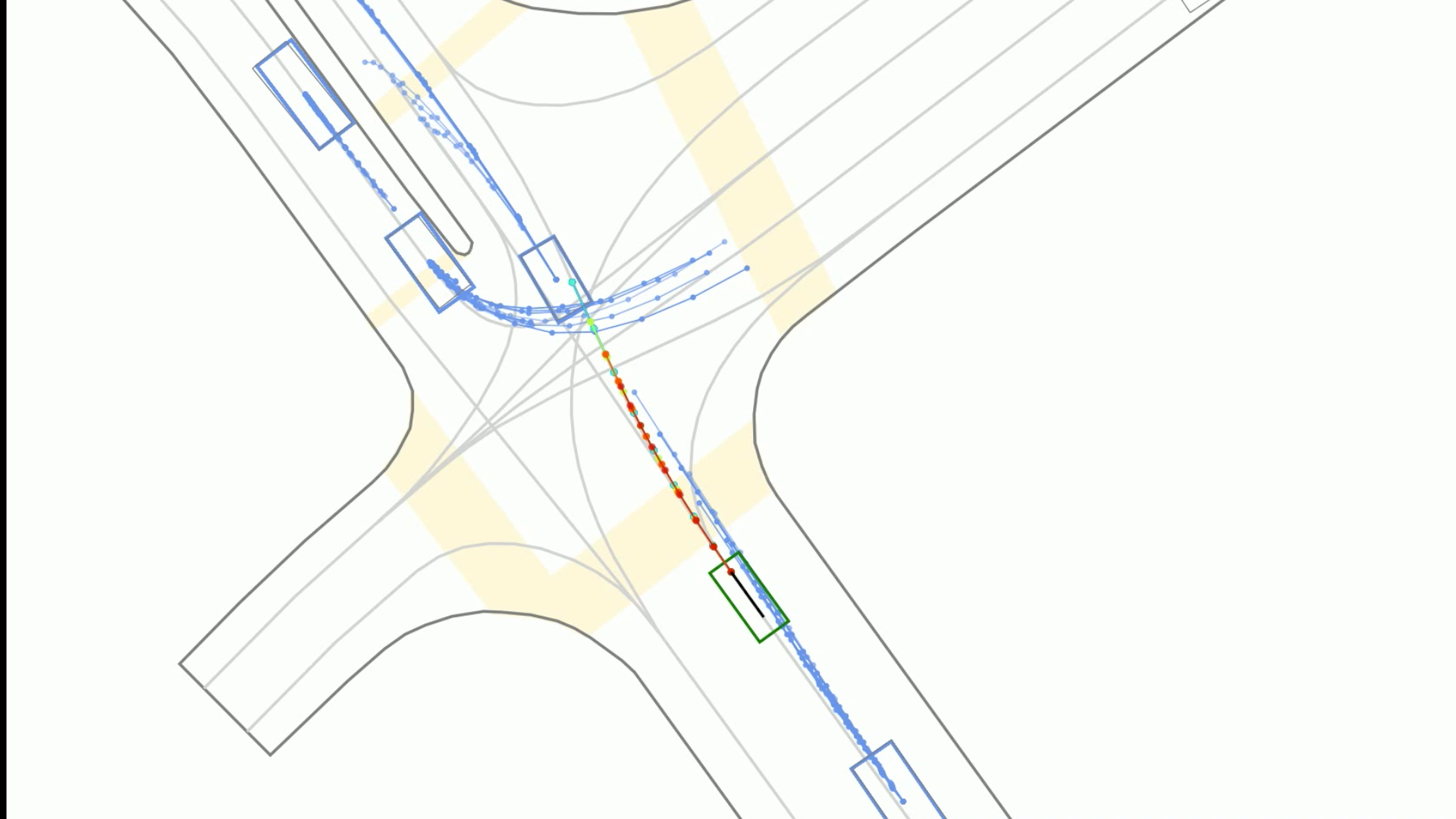}} \\

    \end{tabular}
    \caption{\textbf{Contingency planner qualitative results} in closed-loop simulation. The green bounding box is the SDV. The immediate action (1s) is shown in black starting from the rear axle of the SDV. The contingent trajectories planned for each possible future scenario are shown in distinct colors.}
    \label{fig:qualitative_contingency}
    
\end{figure*}

\begin{figure*}[t]
    
    \begin{tabular}{@{}c@{\hspace{.1em}}c@{\hspace{.5em}}c}
        \textbf{} &
        \textbf{Scenario 1} &
        \textbf{Scenario 2} \\

        \rotatebox[origin=c]{90}{\textbf{\ourmodel{}}} &
        \raisebox{-0.5\height}{\includegraphics[trim={1cm .33cm 1cm .33cm},clip, width=0.45\textwidth]{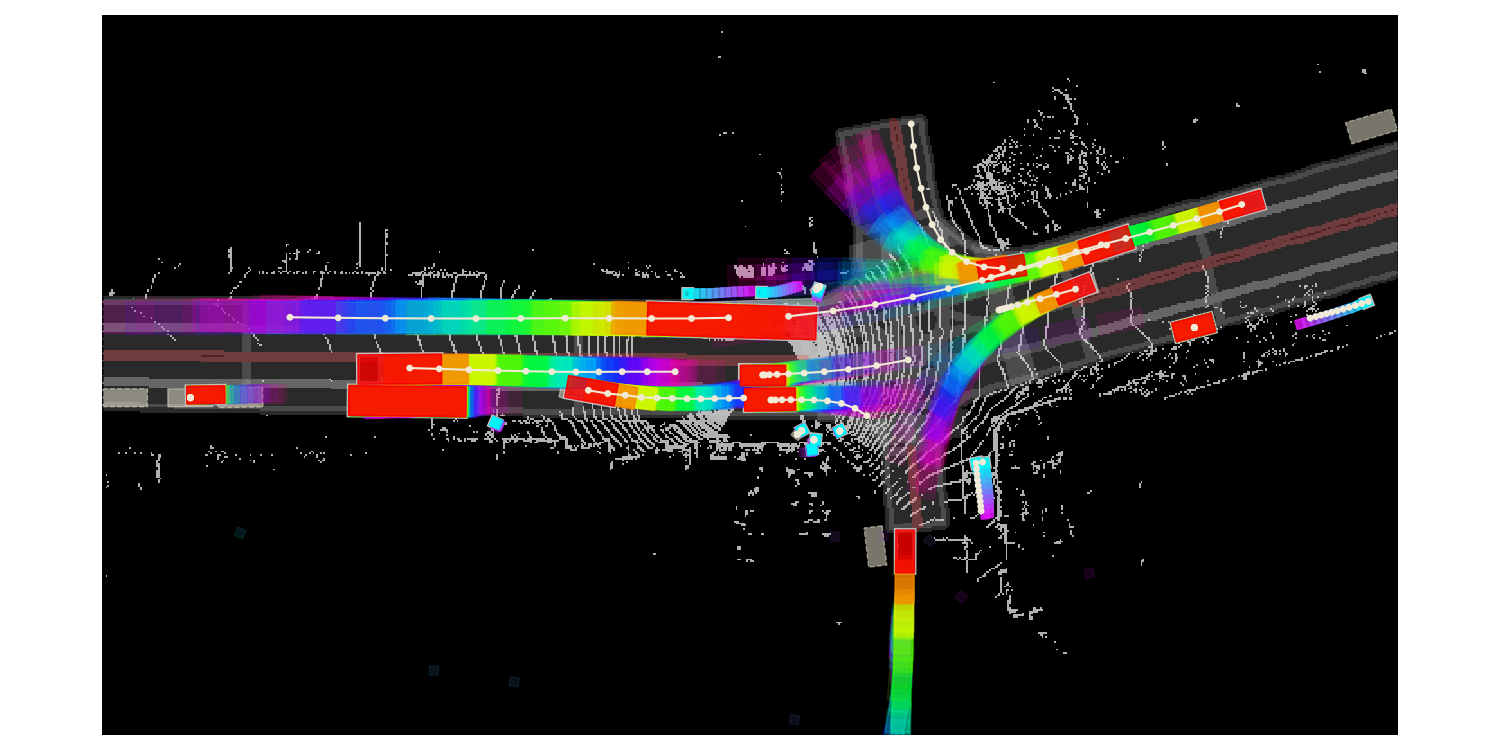}} &
        \raisebox{-0.5\height}{\includegraphics[trim={1cm .33cm 1cm .33cm},clip, width=0.45\textwidth]{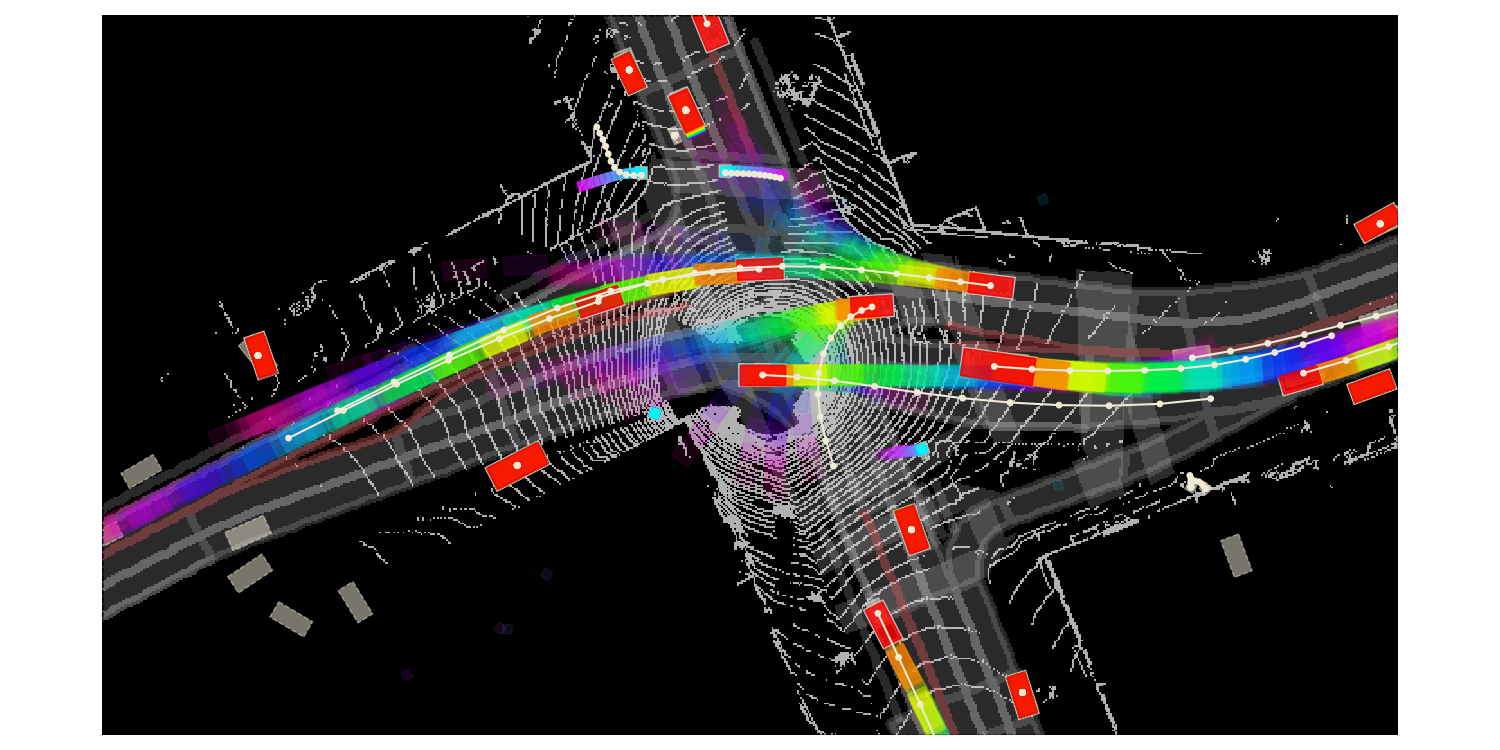}} \\
        \rotatebox[origin=c]{90}{\textbf{MultiPath}} &
        \raisebox{-0.5\height}{\includegraphics[trim={1cm .33cm 1cm .33cm},clip, width=0.45\textwidth]{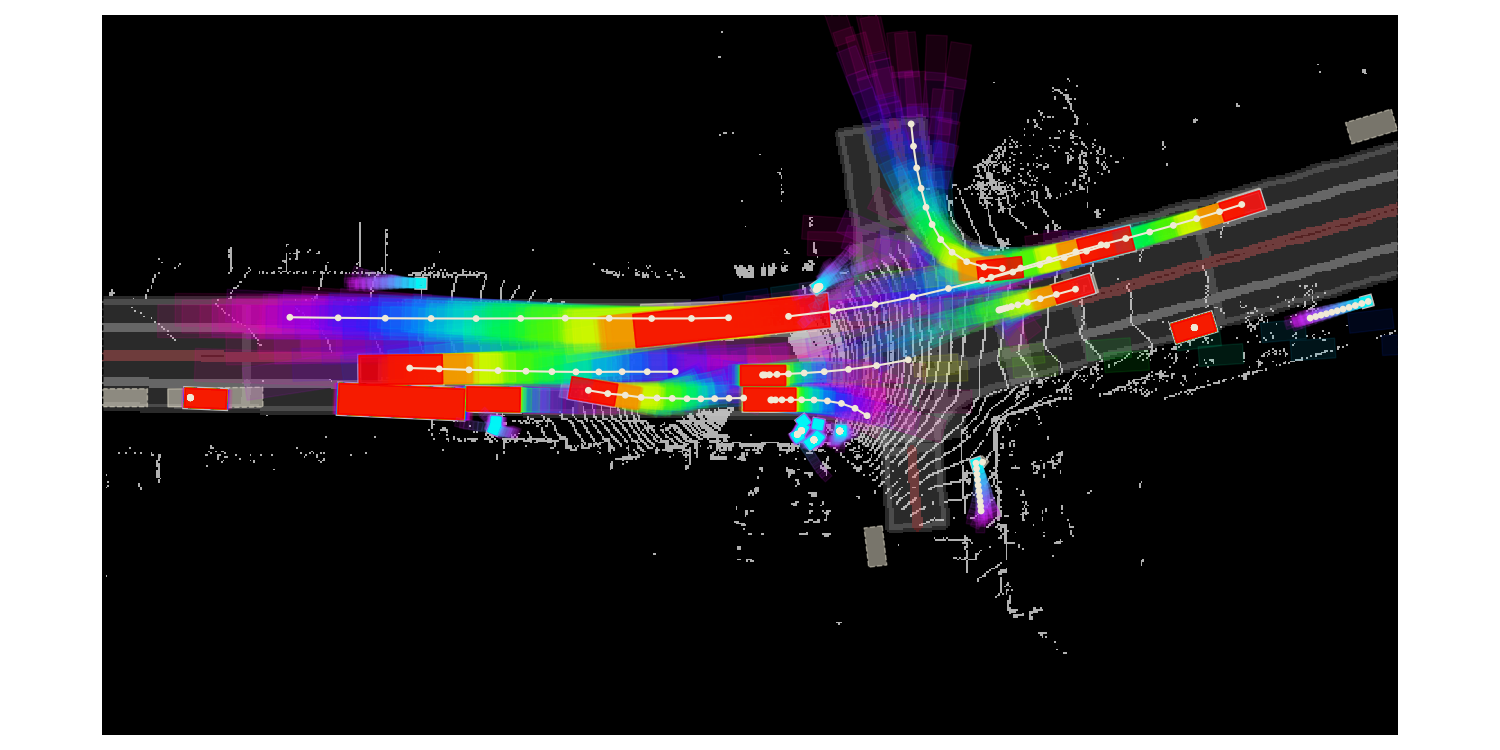}} &
        \raisebox{-0.5\height}{\includegraphics[trim={1cm .33cm 1cm .33cm},clip, width=0.45\textwidth]{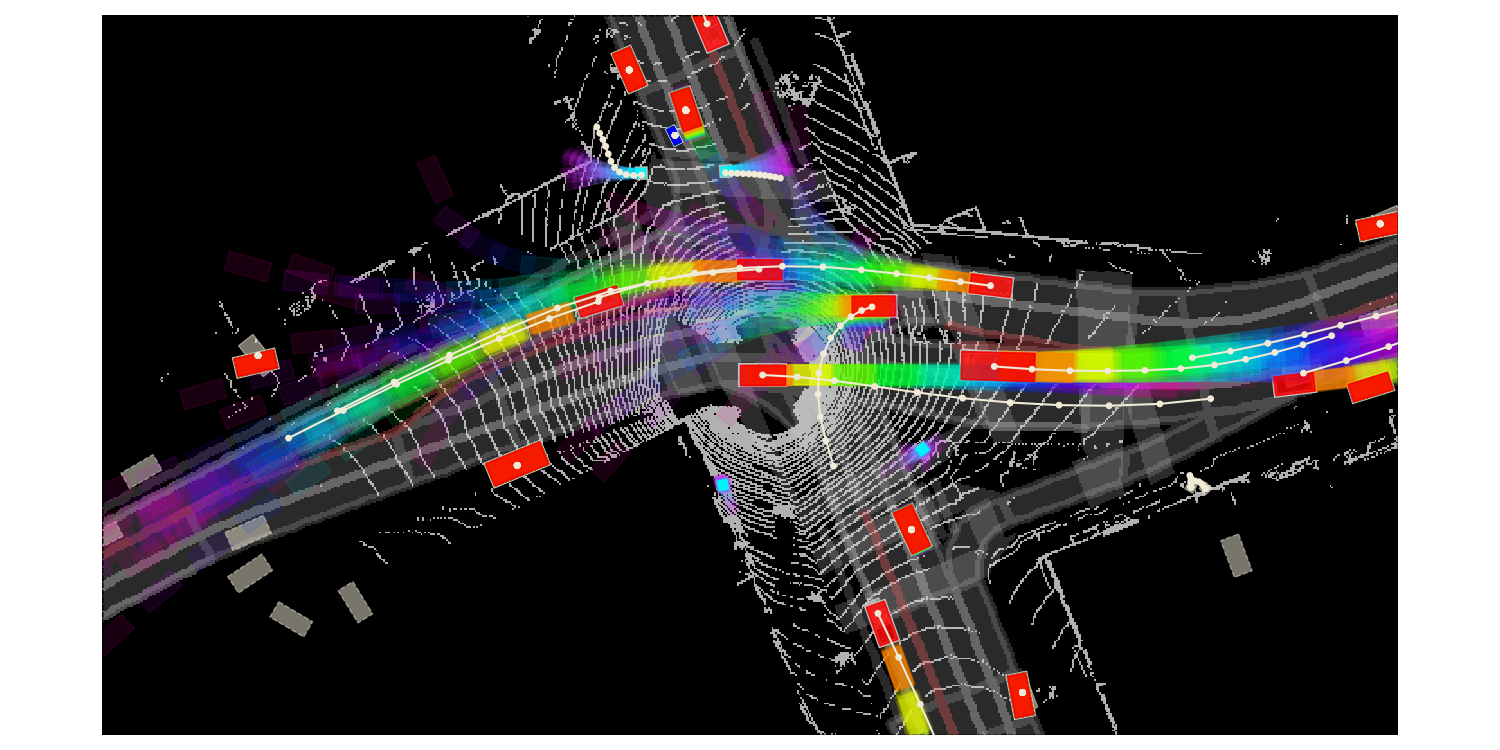}} \\
        \rotatebox[origin=c]{90}{\textbf{ESP}} &
        \raisebox{-0.5\height}{\includegraphics[trim={1cm .33cm 1cm .33cm},clip, width=0.45\textwidth]{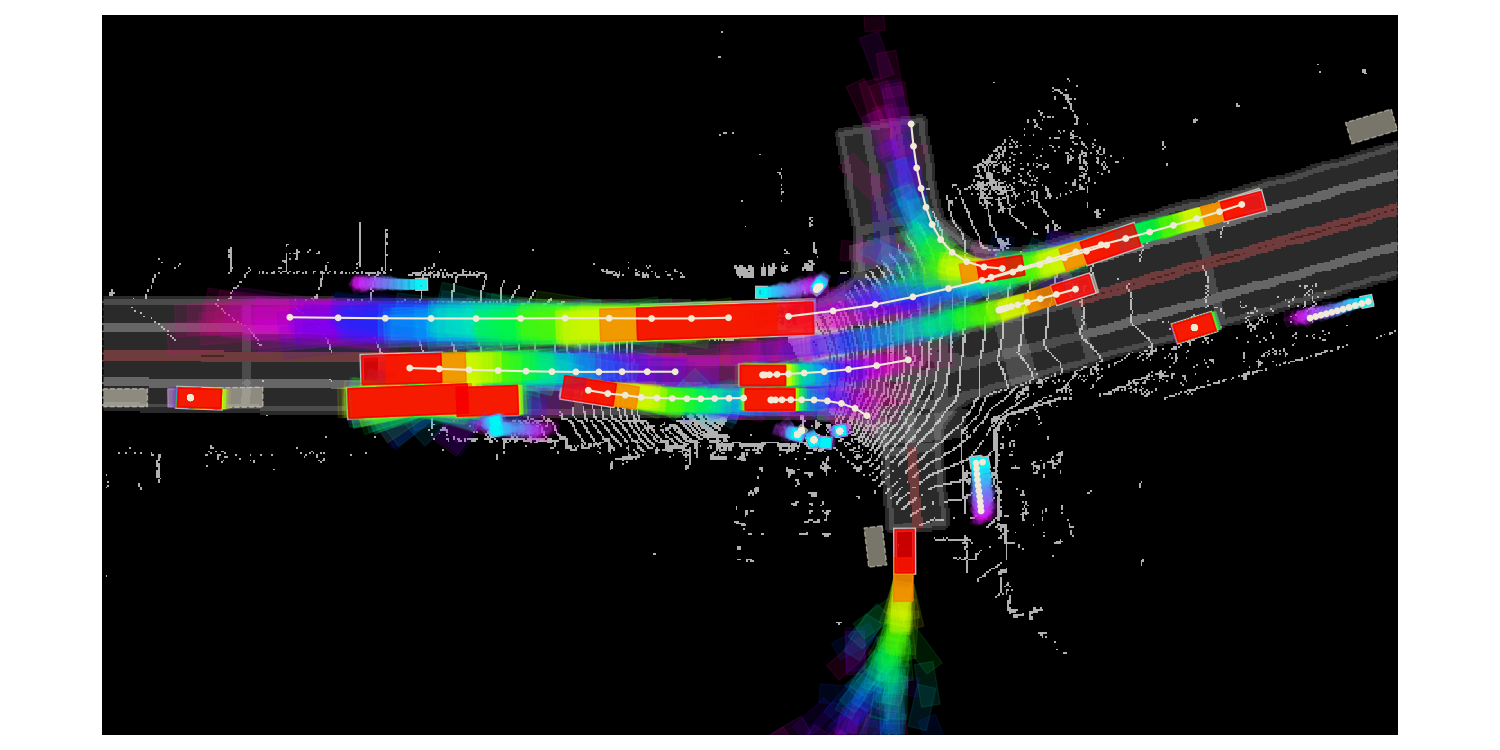}} &
        \raisebox{-0.5\height}{\includegraphics[trim={1cm .33cm 1cm .33cm},clip, width=0.45\textwidth]{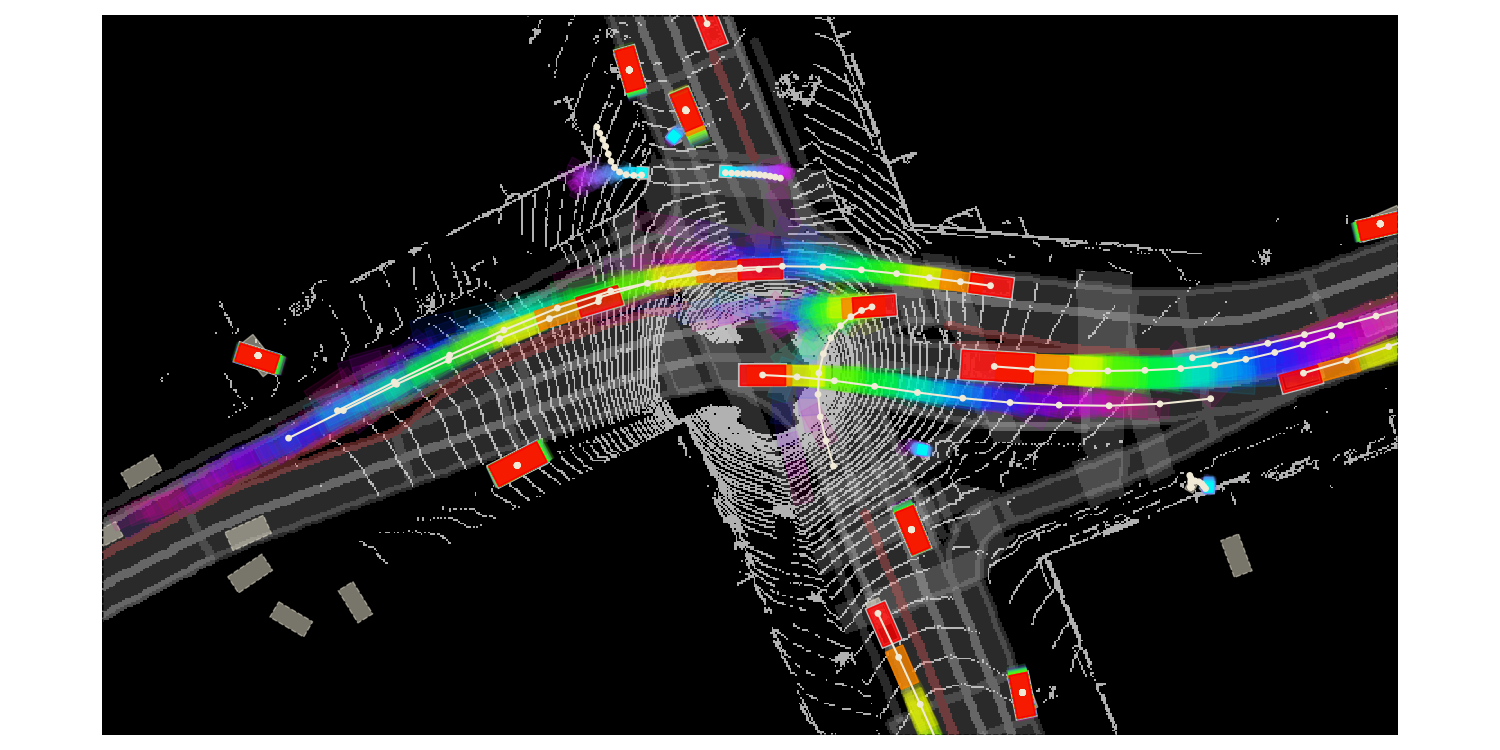}} \\
        \rotatebox[origin=c]{90}{\textbf{ILVM}} &
        \raisebox{-0.5\height}{\includegraphics[trim={1cm .33cm 1cm .33cm},clip, width=0.45\textwidth]{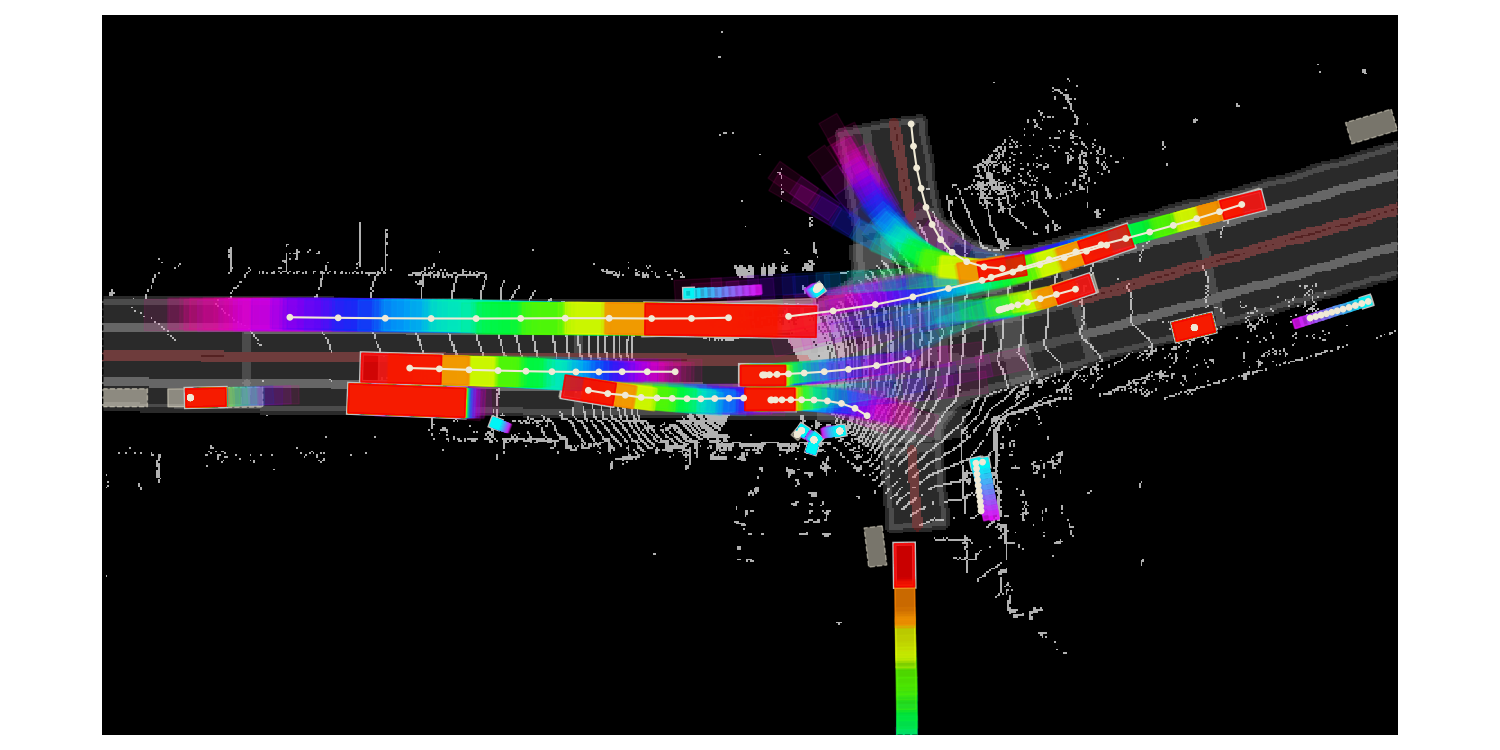}} &
        \raisebox{-0.5\height}{\includegraphics[trim={1cm .33cm 1cm .33cm},clip, width=0.45\textwidth]{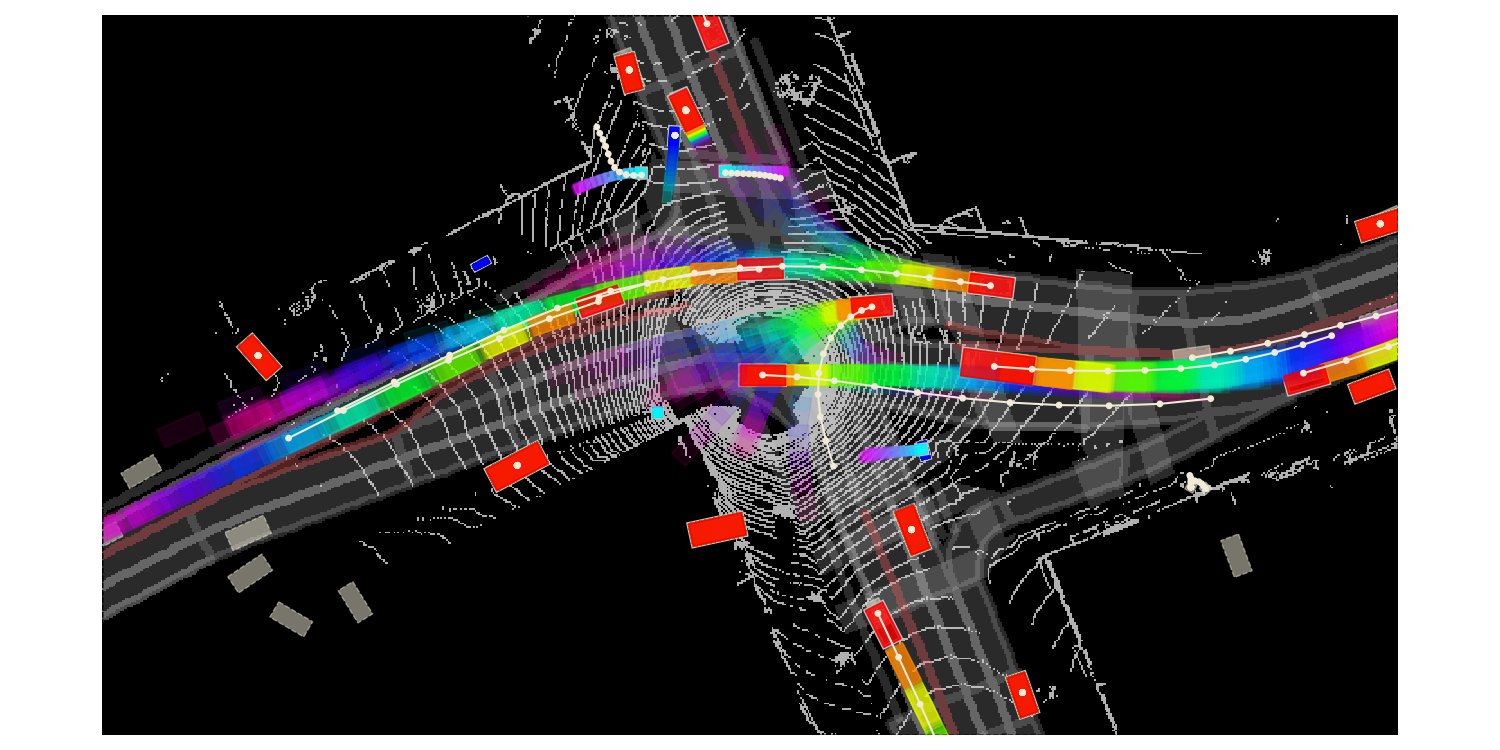}} \\

    \end{tabular}
    \caption{
    \textbf{Motion forecasting visualizations}.
    We blend 15 future scenarios with transparency (we assume equal probability for visualization purposes). Time is encoded in the rainbow color map ranging from red (0s) to pink (5s). 
    This can be seen as a sample-based characterization of the per-actor marginal distributions.}
    \label{fig:qualitative_1}
    
\end{figure*}

\begin{figure*}[t]
    
    \begin{tabular}{@{}c@{\hspace{.1em}}c@{\hspace{.5em}}c}
        \textbf{} &
        \textbf{Scenario 1} &
        \textbf{Scenario 2} \\

        \rotatebox[origin=c]{90}{\textbf{\ourmodel{}}} &
        \raisebox{-0.5\height}{\includegraphics[trim={1cm .33cm 1cm .33cm},clip, width=0.45\textwidth]{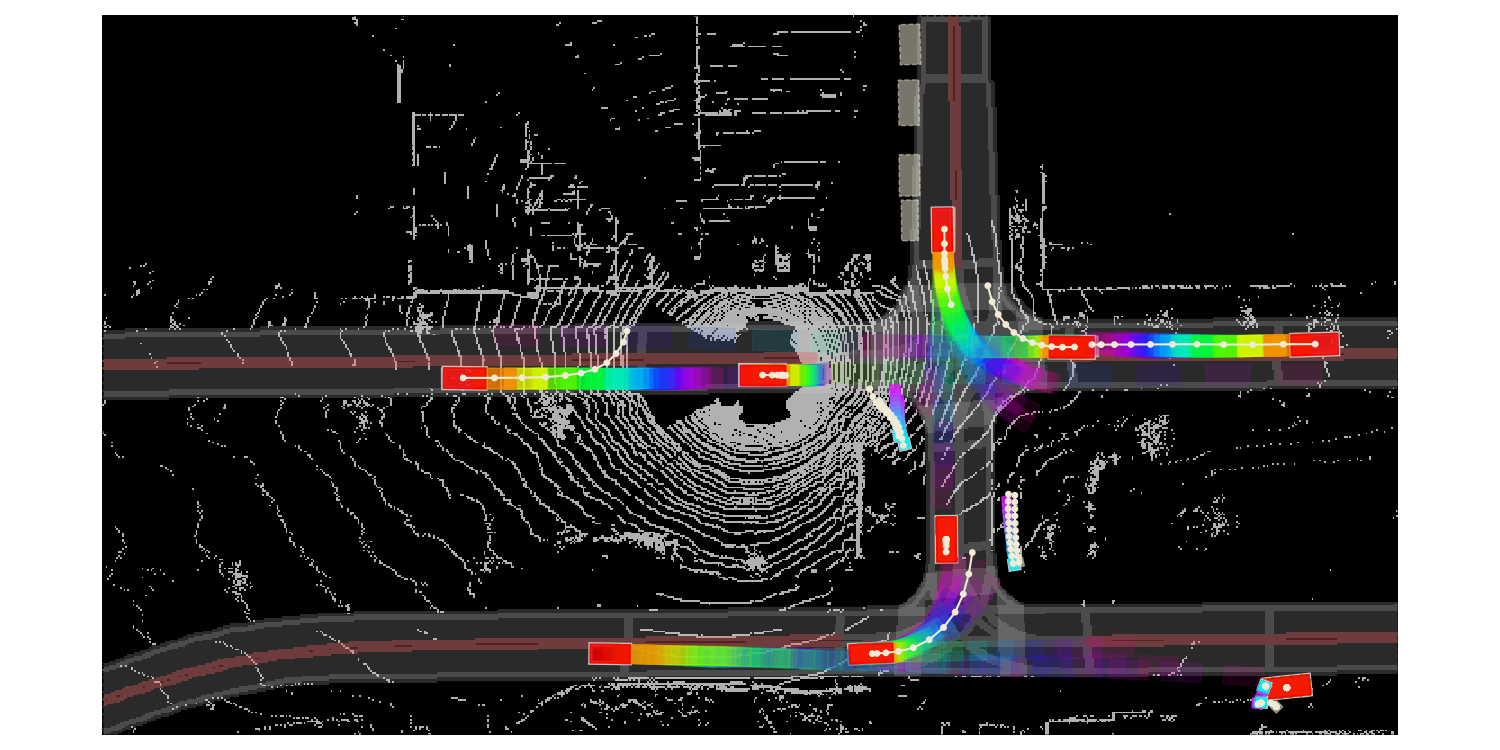}} &
        \raisebox{-0.5\height}{\includegraphics[trim={1cm .33cm 1cm .33cm},clip, width=0.45\textwidth]{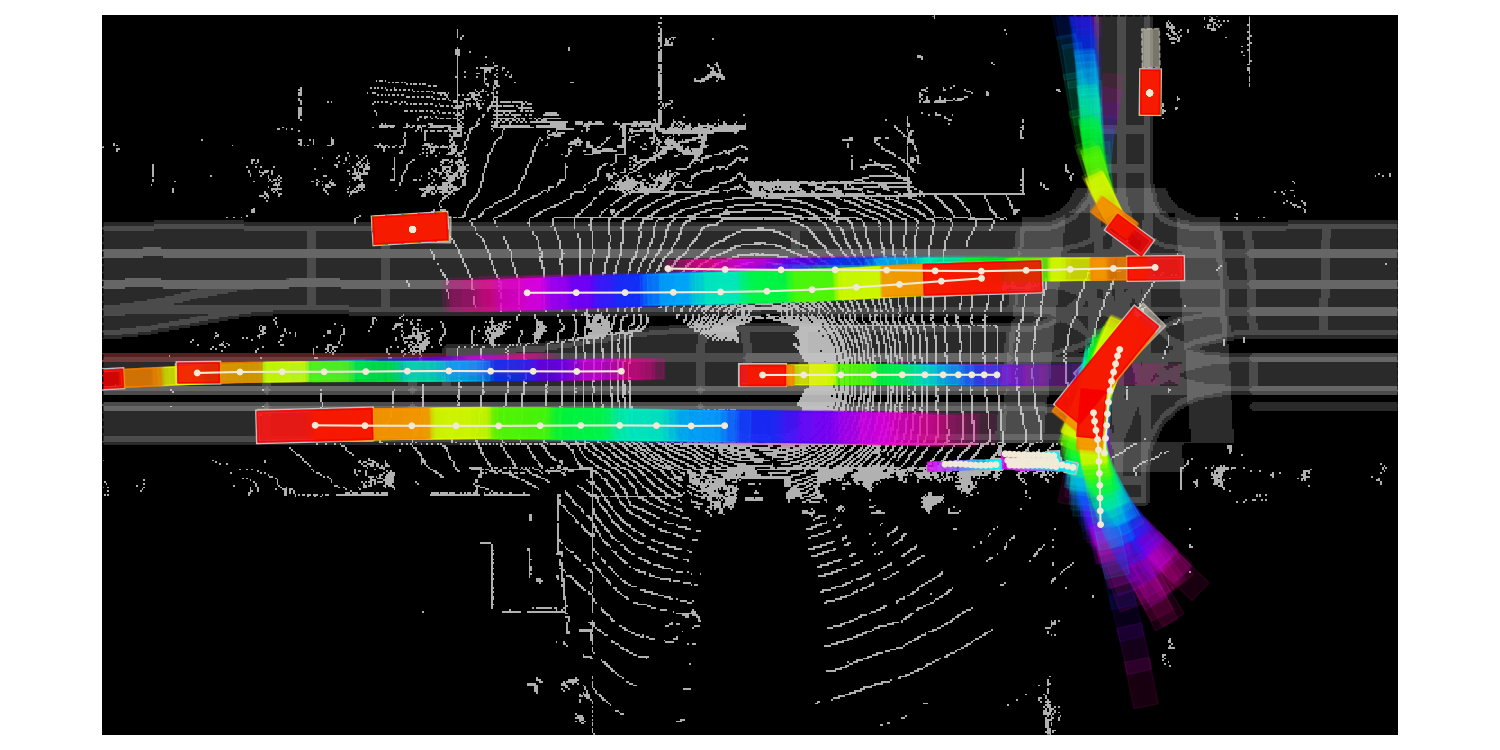}} \\
        \rotatebox[origin=c]{90}{\textbf{MultiPath}} &
        \raisebox{-0.5\height}{\includegraphics[trim={1cm .33cm 1cm .33cm},clip, width=0.45\textwidth]{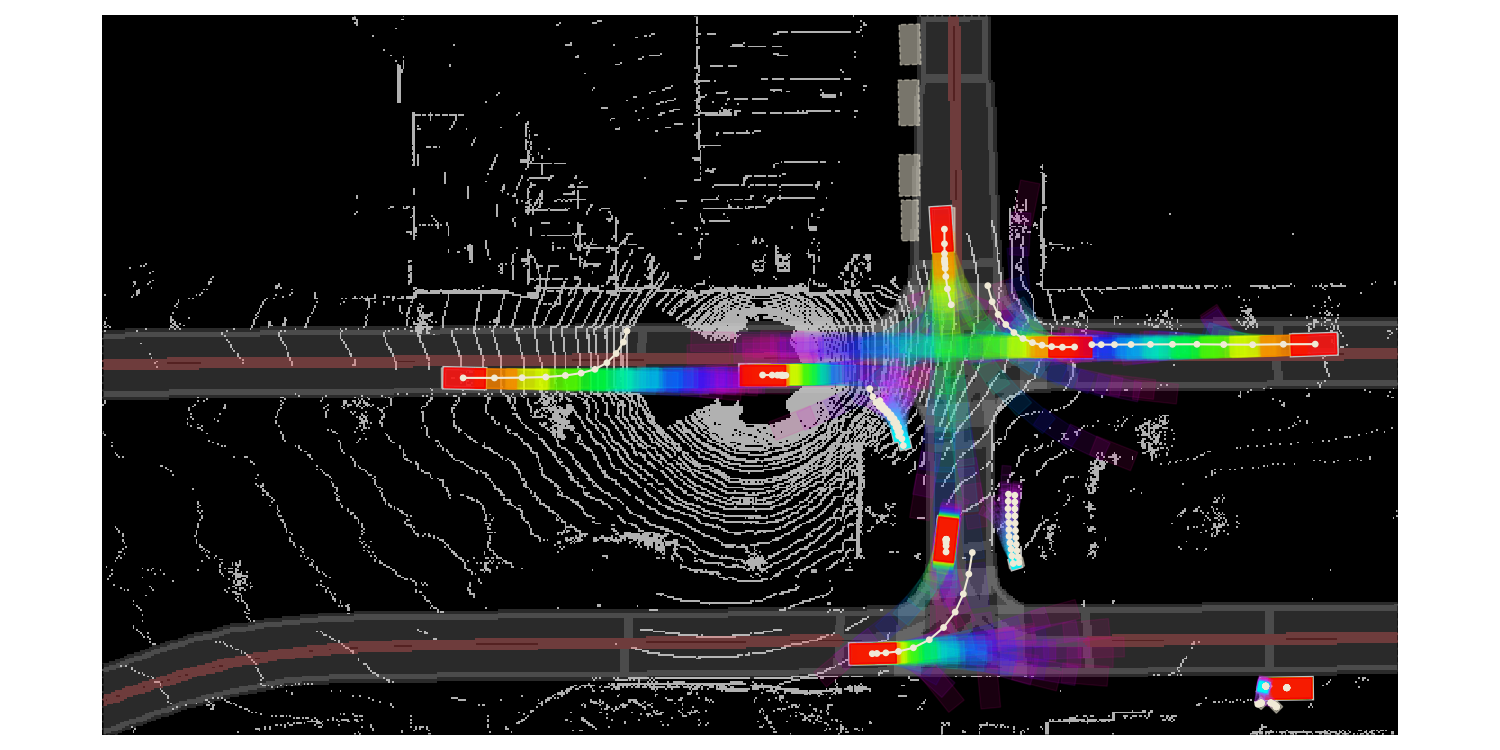}} &
        \raisebox{-0.5\height}{\includegraphics[trim={1cm .33cm 1cm .33cm},clip, width=0.45\textwidth]{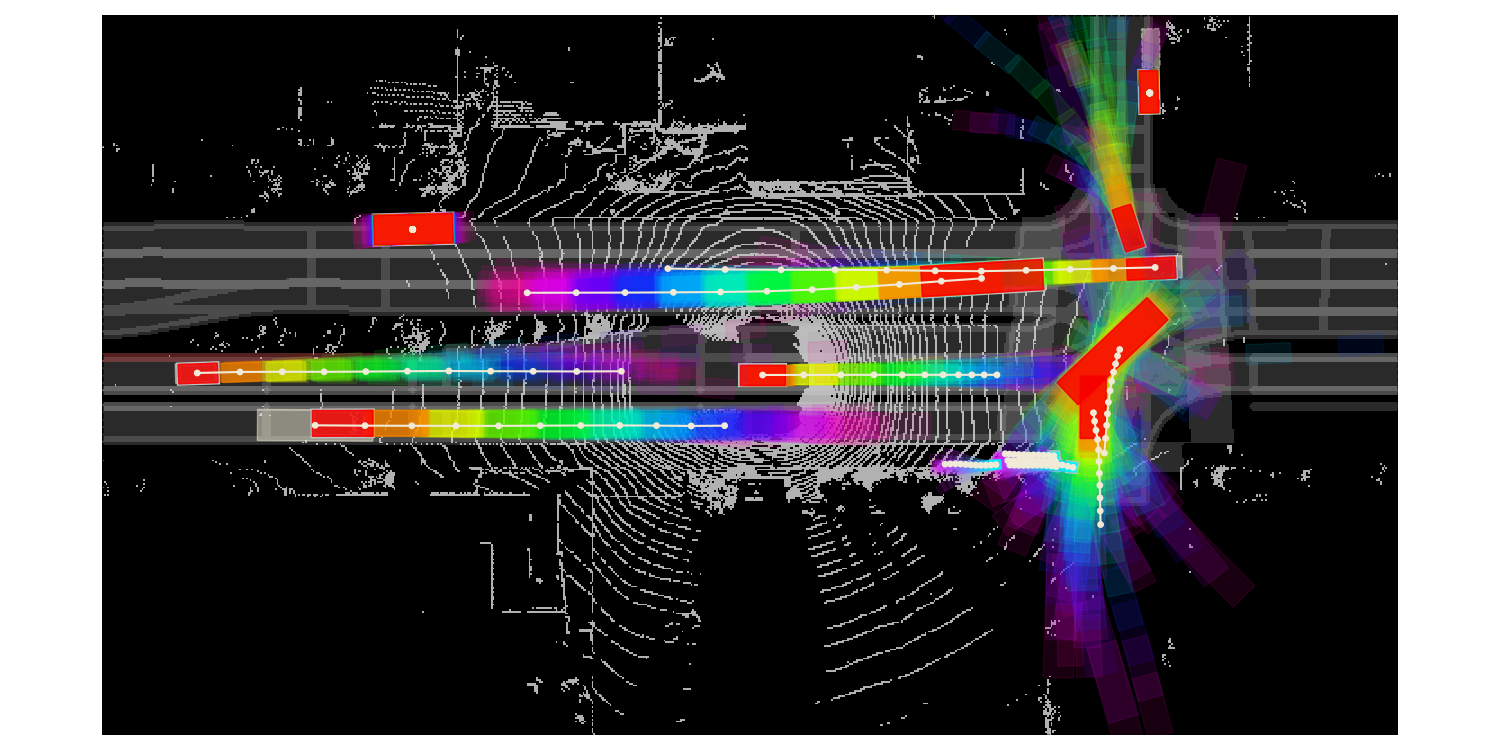}} \\
        \rotatebox[origin=c]{90}{\textbf{ESP}} &
        \raisebox{-0.5\height}{\includegraphics[trim={1cm .33cm 1cm .33cm},clip, width=0.45\textwidth]{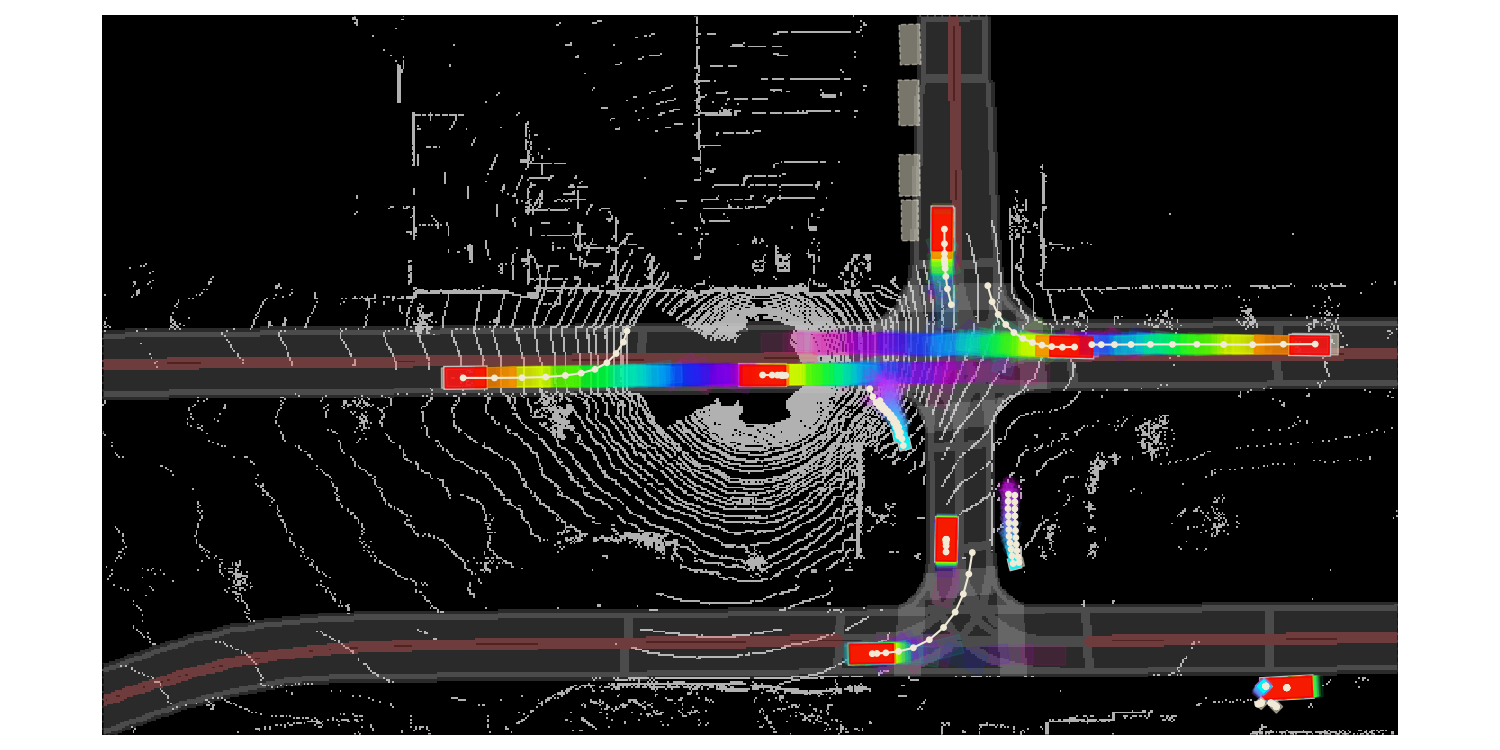}} &
        \raisebox{-0.5\height}{\includegraphics[trim={1cm .33cm 1cm .33cm},clip, width=0.45\textwidth]{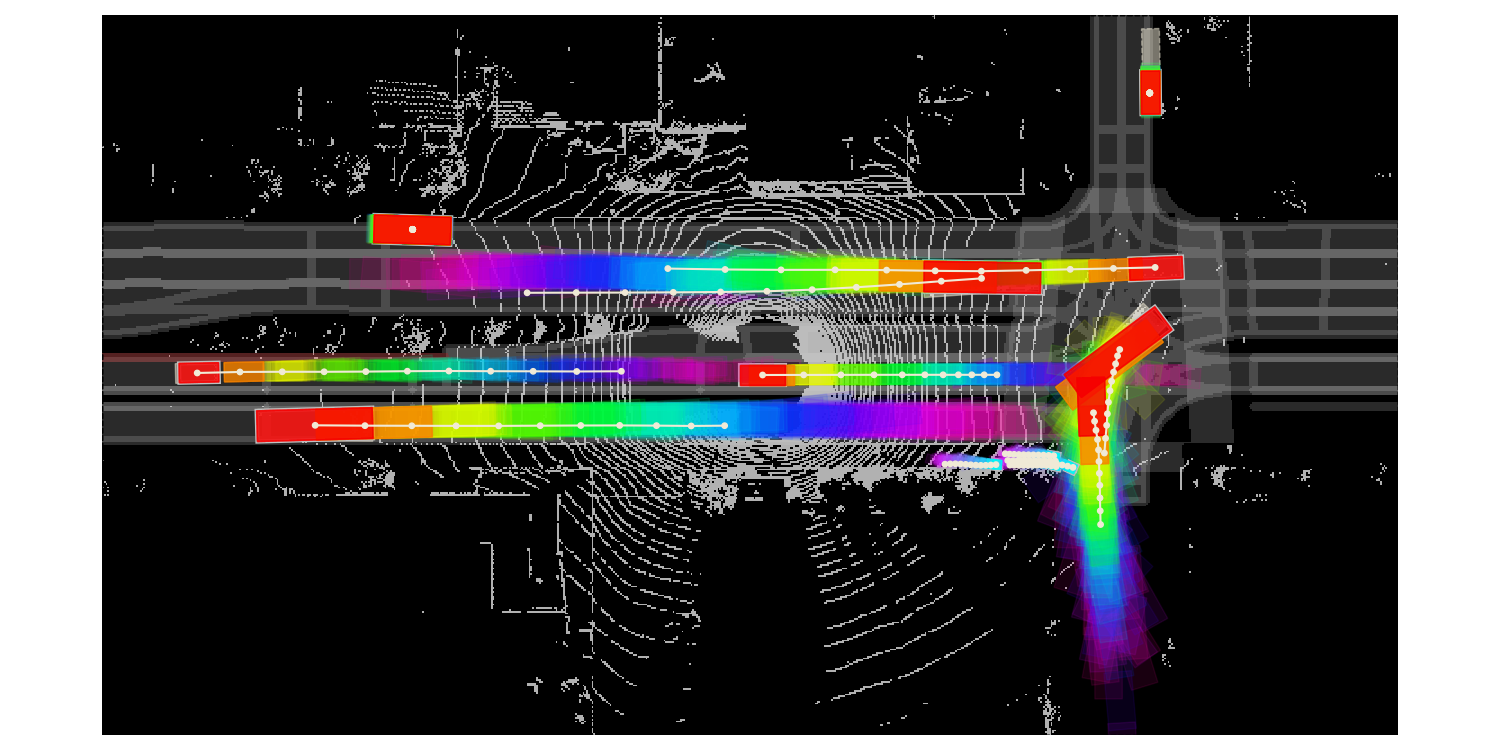}} \\
        \rotatebox[origin=c]{90}{\textbf{ILVM}} &
        \raisebox{-0.5\height}{\includegraphics[trim={1cm .33cm 1cm .33cm},clip, width=0.45\textwidth]{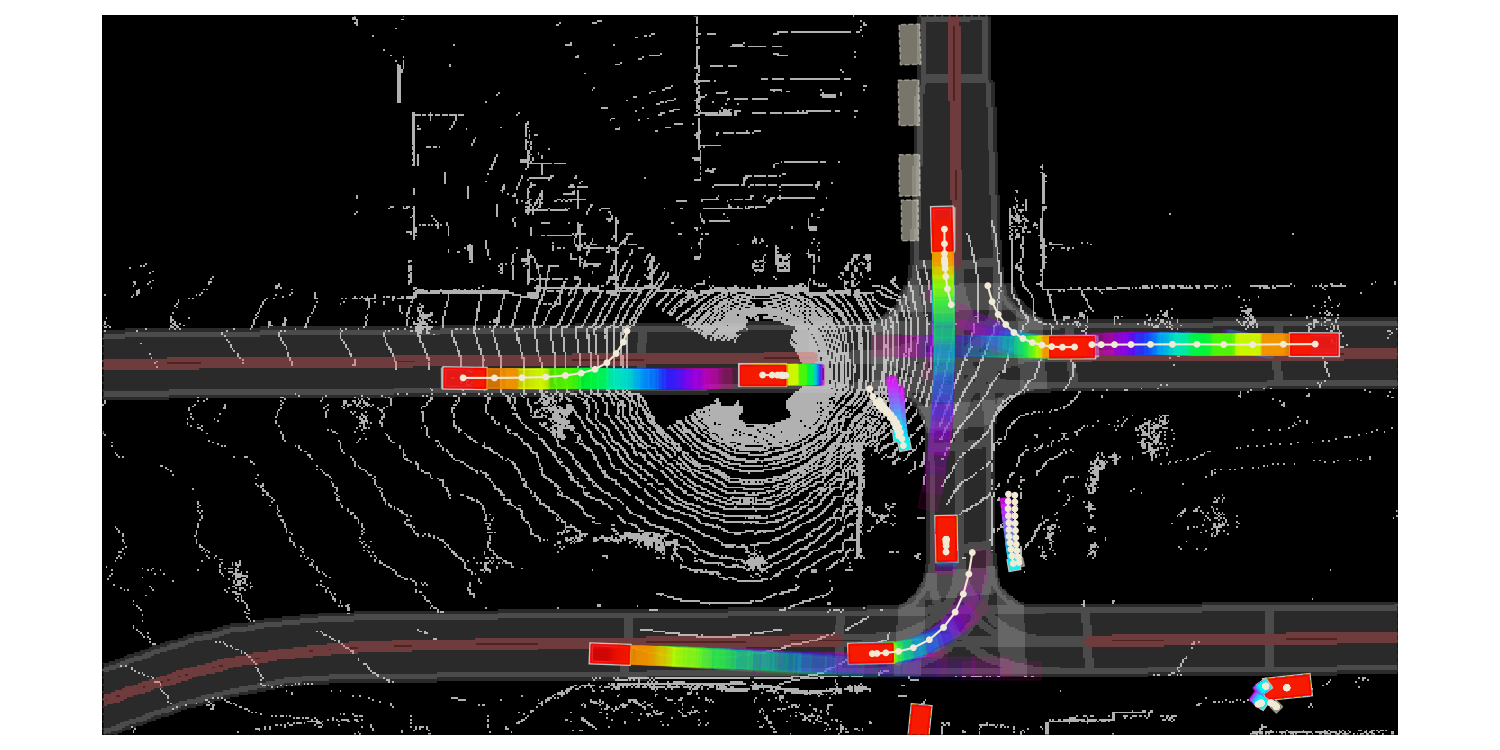}} &
        \raisebox{-0.5\height}{\includegraphics[trim={1cm .33cm 1cm .33cm},clip, width=0.45\textwidth]{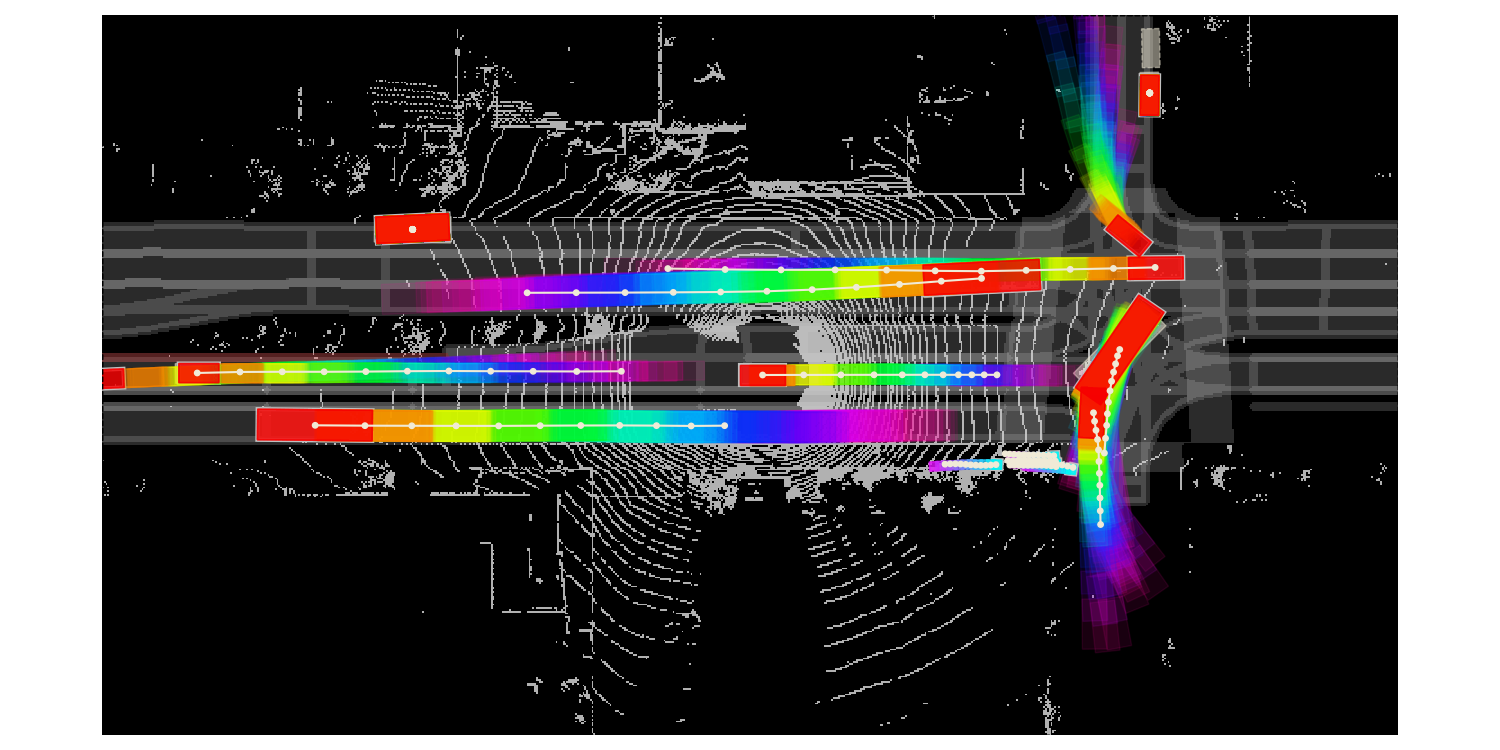}} \\

    \end{tabular}
    \caption{
        \textbf{More motion forecasting visualizations}
    }
    \label{fig:qualitative_2}
    
\end{figure*}

\paragraph{Contingency planner visualizations:} Fig~\ref{fig:qualitative_contingency} contains visualizations for interactive situations that showcase why we need contingency planning.

In Scenario 1, the SDV must decide to go before or after the left turning vehicle at the intersection, which plans to merge into the SDV lane. Since the intersection is crowded and we do not know the exact moment the other vehicle is going to go through, our predictions are diverse in terms of the other actor's speed, planning corresponding safe plans for the distinct futures while keeping a comfortable velocity.

Scenarios 2 and 6 showcase scenarios where the vehicle at the other side of the junction is planning to perform an unprotected left-turn, and the model are not sure whether it will yield to the SDV or not. The SDV plans a cautious immediate action that will allow it to decide to either go or brake when the action from the other actor is more clear.

Scenario 3 shows that when we are following an actor in the right-most lane, we plan slower trajectories in case we need to brake if the actor decides to turn right, and thus slow down.

In Scenario 4, the SDV is taking an unprotected left turn in high moving traffic and considers multiple speed profiles to account for the multiple futures of the incoming traffic.

Finally, Scenario 5 showcases a narrow passage due to parked vehicles, where another actor might either aggressively nudge the parked cars or gently progress. The SDV does not immediately hard-brake because it can plan a safe immediate action that is comfortable and allows it to postpone the decision to whenever more evidence is available or it becomes unsafe to progress.

\paragraph{Motion forecasting Sample Diversity and Quality:}

From Scenario 1 in Fig~\ref{fig:qualitative_1}, we can see that \ourmodel{} shows a diverse range of modalities, and most strongly predicts the forwards acceleration of the SDV (located in the center of the image, next to the curb).

From Scenario 2 in Fig~\ref{fig:qualitative_1}, we see that the left turning vehicle in the center and right turning vehicle about it has a wide range of expected turn modalities. MultiPath demonstrates a wide spread of expected behaviours - however, many in the top left enter the curb, and the left turn on the agent driving from the right is not represented. ESP does the best job in fitting the SDV trajectory, at the cost of diversity.

From Scenario 1 in Fig~\ref{fig:qualitative_2}, we can see \ourmodel{} and MultiPath predict multiple modalities at both intersections, whereas ESP and ILVM trajectories fit one mode predominantly at each intersection.

From Scenario 2 in Fig~\ref{fig:qualitative_2}, we mainly see that MultiPath struggles here in making realistic predictions, with many entering the curb. On the other hand, \ourmodel{} demonstrates diversity mostly in speed and the trajectory of the bus, which is in the path of the SDV and thus is important to model diversely.

\end{document}